%% file: main.tex
\documentclass[journal]{IEEEtran}

\usepackage{amsmath,amsfonts}
\usepackage{algorithmic}
\usepackage{algorithm}
\usepackage[caption=false,font=normalsize,labelfont=sf,textfont=sf]{subfig}
\usepackage{textcomp}
\usepackage{stfloats}
\usepackage{url}
\usepackage{verbatim}
\usepackage{graphicx}
\usepackage{cite}
\usepackage{xcolor}
\usepackage{balance} 
\usepackage{silence} 

\usepackage[
    colorlinks=true,
    linkcolor=blue,
    citecolor=blue,
    urlcolor=black
]{hyperref}

\usepackage{orcidlink} 

\usepackage{etoolbox}
\AtBeginDocument{%
    \renewcommand{\cite}[1]{\textcolor{blue}{[\citenum{#1}]}}
}

\usepackage{csquotes}      
\usepackage[capitalise]{cleveref}	
\crefname{figure}{Fig.}{Figs.} 
\Crefname{figure}{Fig.}{Figs.} 
\usepackage{color,soul}
\usepackage{amssymb}   
\usepackage{pifont}         
\newcommand{\cmark}{\ding{51}}  
\newcommand{\xmark}{\ding{55}}  
\newcommand{\dmark}{\ding{108}}

\usepackage{array}
\usepackage{booktabs}
\usepackage{vcell}
\usepackage{tabularx} 
\usepackage{fontsize} 
\usepackage{makecell}
\usepackage{multirow}  
\usepackage{adjustbox} 

\definecolor{mydarkgreen}{RGB}{0,200,0}

\usepackage{tabularray}
\usepackage{acro}[=v2]   
\usepackage{enumitem}
\usepackage{mfirstuc}  
\usepackage{microtype}  

\setlist{nolistsep}    

\DeclareAcroListStyle{mydesc}{list}{
  list = description ,
}

\acsetup{
  list-style = mydesc
}

\setlist[description]{style=standard, leftmargin=2.5cm, labelsep=0pt, labelwidth=\dimexpr2.5cm-\labelsep}

\input{abbr}

\begin{document}
\bstctlcite{IEEEexample:BSTcontrol}

\title{The Role of Generative Artificial Intelligence in Internet of Electric Vehicles}

\author{Hanwen Zhang\raisebox{0.5ex}{\orcidlink{0000-0002-6295-4753}},~\IEEEmembership{Student Member,~IEEE},~Dusit Niyato\raisebox{0.5ex}{\orcidlink{0000-0002-7442-7416}},~\IEEEmembership{Fellow,~IEEE,}~Wei Zhang\raisebox{0.5ex}{\orcidlink{0000-0002-2644-2582}},~\IEEEmembership{Member,~IEEE,}
\\
~Changyuan Zhao\raisebox{0.5ex}{\orcidlink{0000-0001-9187-9572}},~Hongyang Du\raisebox{0.5ex}{\orcidlink{0000-0002-8220-6525}},~Abbas Jamalipour\raisebox{0.5ex}{\orcidlink{0000-0002-1807-7220}},~\IEEEmembership{Fellow,~IEEE,}~\\Sumei Sun\raisebox{0.5ex}{\orcidlink{0000-0002-1701-8122}},~\IEEEmembership{Fellow,~IEEE,}~Yiyang Pei\raisebox{0.5ex}{\orcidlink{0000-0002-9915-8697}},~\IEEEmembership{Senior Member,~IEEE}

\thanks{Manuscript received January 1, 2000; revised January 1, 2000. This research/project is supported by the National Research Foundation, Singapore and Infocomm Media Development Authority under its Future Communications Research \& Development Programme (Grant FCP-SIT-TG-2022-007), A*STAR under its MTC Programmatic (Award M23L9b0052), MTC Individual Research Grants (IRG) (Award M23M6c0113), the Ministry of Education, Singapore, under the Academic Research Tier 1 Grant (Grant ID: GMS 693), and SIT’s Ignition Grant (STEM) (Grant ID: IG (S) 2/2023 – 792). (\textit{Corresponding author: Wei Zhang})}
\thanks{Hanwen Zhang is with both the College of Computing and Data Science, Nanyang Technological University, Singapore 639798, and the Information and Communications Technology Cluster, Singapore Institute of Technology, Singapore 138683 (e-mail: hanwen001@e.ntu.edu.sg and hanwen.zhang@singaporetech.edu.sg)}
\thanks{Wei Zhang, Sumei Sun, and Yiyang Pei are with the Information and Communications Technology Cluster, Singapore Institute of Technology, Singapore 138683 (e-mail: \{wei.zhang, sumei.sun, yiyang.pei\}@singaporetech.edu.sg).}
\thanks{Dusit Niyato and Changyuan Zhao are with the College of Computing and Data Science, Nanyang Technological University, Singapore 639798 (e-mail: dniyato@ntu.edu.sg and zhao0441@e.ntu.edu.sg).}
\thanks{Hongyang Du is with the Department of Electrical and Electronic Engineering, University of Hong Kong, Pok Fu Lam, Hong Kong (e-mail: duhy@eee.hku.hk).}
\thanks{Abbas Jamalipour is with the School of Electrical and Computer Engineering, the University of Sydney, Australia (e-mail: abbas.jamalipour@sydney.edu.au).}
}
\maketitle

\begin{abstract}

With the advancements of \ac{GenAI} models, their capabilities are expanding significantly beyond content generation and the models are increasingly being used across diverse applications. Particularly, \ac{GenAI} shows great potential in addressing challenges in the \ac{EV} ecosystem ranging from charging management to cyber-attack prevention. In this paper, we specifically consider \ac{IoEV} and we categorize \ac{GenAI} for \ac{IoEV} into four different layers namely, EV's battery layer, individual \ac{EV} layer, smart grid layer, and security layer. We introduce various \ac{GenAI} techniques used in each layer of \ac{IoEV} applications. Subsequently, public datasets available for training the \ac{GenAI} models are summarized. Finally, we provide recommendations for future directions. 
This survey not only categorizes the applications of \ac{GenAI} in IoEV across different layers but also serves as a valuable resource for researchers and practitioners by highlighting the design and implementation challenges within each layer. Furthermore, it provides a roadmap for future research directions, enabling the development of more robust and efficient IoEV systems through the integration of advanced \ac{GenAI} techniques.
\end{abstract}


\section{Introduction}
\label{sec-intro}
\IEEEPARstart{E}{lectric} mobility is the future. This is evidenced by the visions and policies of different nations and companies for achieving sustainable mobility. \Ac{EVs} are gaining popularity rapidly. In the US, the sales of \acs{EV} reached 1.2 million just in one quarter in 2023, with nearly 8\% market share \cite{web_BBC1}. The numbers are even bigger in China. In 2023, EV's market share was 34\% \cite{web_yahoofinance}. The trend is not much different in other nations, e.g., the sales of traditional cars with \ac{ICEs} will be phased out by 2030 and 100\% cars will be clean energy based after 2040 in Singapore. Altogether, the global market of \ac{EVs} reached 392.4 billion USD in 2023 with a predicted compound annual growth rate of 13.9\% from 2024 to 2032 \cite{web_VANTAGE}. With more and more \ac{EVs} on the road, they naturally form a \ac{IoEV} \cite{zhang2023urban}. It shares a similar spectrum of technologies to \ac{IoT} and offers new features such as connectivity, grid services, predictive maintenance, and traffic management.
\Ac{IoEV} could be regarded as a specialized subset of \ac{IoT} where \ac{EVs}, their charging infrastructure, and smart grid interact seamlessly \cite{IoTJ_Ref3}. This creates a network of interconnected devices/systems, leading to various applications, e.g.,  \acs{EV} routing problem in \ac{IoEV} \cite{IoTJ_Ref2}, smart \acs{EV} charging station scheduling \cite{IoTJ_Ref4}, EV's battery life prediction \cite{IoTJ_Ref6}, and blockchain-based bidirectional energy trading between \ac{EVs} and charging stations \cite{IoTJ_Ref5}. 

Same as most new technologies, \acs{EV} and \ac{IoEV} have their problems and challenges. Making electricity the main source of power brings constraints related to electricity at the same time. The electricity is stored in batteries, which have charging speed and capacity limits, can degrade over time, and may catch fire occasionally. Those constraints affect \acs{EV} operations in various aspects, e.g., charging scheduling and battery health monitoring. Furthermore, the impact goes beyond individual \acs{EV} for \ac{IoEV} with increased coordination and intelligence demand. One impact is on the power grid, where a significant amount of new electricity load from EVs shall not stress the grid much and hurt the grid's stability. This requires grid-level intelligence such as supply-demand forecasting and matching with the support of \ac{IoEV} scheduling and \ac{V2G} services.

The problems and challenges have drawn attention from the research communities and industries. Existing efforts can be clustered into application level and technology level. The counterparts of \ac{EV} and \ac{IoEV} are traditional cars with \ac{ICEs}. \Ac{IoEV} is a subset of \ac{IoV}, with high relevance. Research studies for these applications have the potential of being transferred to \ac{IoEV}, but this comes with certain constraints and potentially substantial costs for some tasks due to the new challenges of \ac{IoEV}. For example, charging typically takes hours and cannot be modeled as an instant event, and electricity prices may change hourly or more frequently. As such, the \ac{EV} charging management system shall schedule and predict the \ac{EV} loads to minimize cost and avoid overloading the power grid, and charging station installation and operation shall be optimized to promote charging service availability.

Technology-wise, there are solutions specialized to \ac{IoEV}, with or without using \ac{ML}. Non-ML solutions often use model-based statistical methods for various \ac{IoEV} related tasks, e.g., user behavior understanding \cite{ Ref_normal_distribution1} and charging scheduling \cite{Ref_NonAI2}. A common problem of these methods is that the real system dynamics are modeled in a highly summarized way, e.g., mean and variance, and formulated with simplification to make the optimization tractable. As a result, they are often impractical and fail to offer sufficient accuracy and effectiveness in modeling and charging scheduling. ML-based solutions are becoming the trend. \Ac{ML} models learn from data and continuously improve with more data available. For example, the \ac{LSTM} is often used in the prediction of \ac{EV} load \cite{Ref5} and voltage changes of battery \cite{Ref3} because of its superior capability of handling long-term dependencies in sequential data, yet the prediction accuracy decreases when uncertainty in the real systems increases. 

\begin{table}
	\centering
        \caption{List of abbreviations.} 
        \label{tab:list_of_abbr} 
	\begin{tabular}{l||l}
		\toprule
		\Xhline{1pt} 
		\multicolumn{1}{c||}{\textbf{Abbreviation}} & \multicolumn{1}{c}{\textbf{Description}}  \\ 
		\Xhline{0.75pt} 
		\hline
		AE & Autoencoder \\ \hline
		AM & DRL with general attention model \\ \hline
		ANN & Artificial neural network \\ \hline
		AutoML & Automated machine learning \\ \hline
		BMS & Battery management system \\ \hline
		CAN & Controller area network \\ \hline
		CNN & Convolutional neural network \\ \hline
		DBN & Deep belief network \\ \hline
		DDIM & Denoising diffusion implicit model \\ \hline
		DDPM & Denoising diffusion probabilistic model \\ \hline
		DNN & Deep neural network \\ \hline
		DoS & Denial of service \\ \hline
		DRL & Deep reinforcement learning \\ \hline
		ESS & Energy storage system \\ \hline
		EV & Electric vehicle \\ \hline
		EVRP & Electric vehicle routing problem \\ \hline
		FDIAs & False data injection attacks \\ \hline
		FGSM & Fast gradient sign method \\ \hline
		FNN & Feed-forward neural network \\ \hline
		GAE & Graph autoencoder \\ \hline
		GenAI & Generative artificial intelligence \\ \hline
		GAN & Generative adversarial network \\ \hline
		GDM & Generative diffusion model \\ \hline
		GMM & Gaussian mixture model \\ \hline
		GPR & Gaussian process regression \\ \hline
		EMS & Energy management system \\ \hline
		IoEV & Internet of electric vehicles \\ \hline
		IoT & Internet of things \\ \hline
		IoV & Internet of vehicle \\ \hline
		KNN & $k$-nearest neighbor \\ \hline
		LLMs & Large language models \\ \hline
		LSTM & Long short-term memory \\ \hline
		MADRL & Multi-agent deep reinforcement learning \\ \hline
		MAE & Mean absolute error \\ \hline
		MARL & Multi-agent reinforcement learning \\ \hline
		ML & Machine learning \\ \hline
		NLP & Natural language processing \\ \hline
		PPO & Proximal policy optimization \\ \hline
		PV & Photovoltaic \\ \hline
		RL & Reinforcement learning \\ \hline
		RNN & Recurrent neural network \\ \hline
		SAC & Soft actor-critic \\ \hline
		SoC & State of charge \\ \hline
		SoH & State of health \\ \hline
		SVM & Support vector machine \\ \hline
		VAE & Variational autoencoder \\ \hline
		VRP & Vehicle routing problem \\ 
		\Xhline{1pt} 
		\bottomrule
	\end{tabular}
\end{table}

\begin{figure*}[ht!]
	\centerline{
            \includegraphics[width=1.0\textwidth]{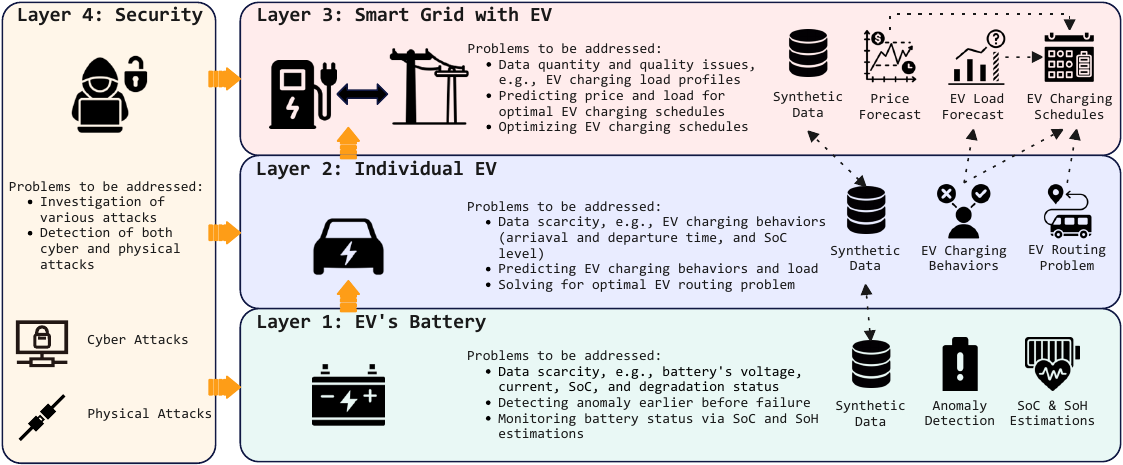}
	}
        \caption{\Ac{GenAI} for \ac{IoEV} applications can be categorized into four layers: 
            Layer 1's problems are anomaly detection, \ac{SoC} estimation, and \ac{SoH} evaluation. 
            Layer 2 primarily focuses on data augmentation in \ac{EV} charging behaviors, prediction of an \ac{EV} load at home, and solving the optimal \ac{EV} routing problem. 
            Layer 3 concentrates on: Forecasting and augmenting the \ac{EV} charging load profiles; Optimizing \ac{EV} charging schedules based on the constraints from either \ac{EV} charging stations or the smart grid; Predicting the electricity price is necessary for \ac{EV} charging station operators. 
            Layer 4 studies various attacks which may be harmful to the \ac{EV} and the charging system. Both cyber and physical attacks need to be studied and detected such as adversarial attacks, false data injection attacks, denial of service attacks, fuzzy attacks, and impersonation attacks.
  }
    \label{Fig1:Layers}
\end{figure*}

In this paper, we propose to use \ac{GenAI} to advance \ac{IoEV} technologies. \Ac{GenAI} techniques applied in \ac{IoEV} mirror the same advancements in \ac{IoT} where a large amount of data from various sources are analyzed and utilized to enhance efficiency, safety, and user experience. The integration of \ac{GenAI} within \ac{IoEV} not only enhances its specific applications but also contributes to the overarching goals of \ac{IoT} by enabling smarter, more autonomous, and connected systems. Moreover, \ac{GenAI} has the potential to address the challenges mentioned above and case studies for certain \ac{IoEV} tasks are available, e.g., charging demand forecasting \cite{Ref21, Ref23} and data augmentation \cite{Ref18, Ref27}. We aim to go beyond case studies and provide a comprehensive discussion of GenAI's roles in the \ac{IoEV} ecosystem. Specifically, we structure the system into four layers as shown in \cref{Fig1:Layers}. The bottom layer is the battery, which is a critical component of an \ac{EV} and brings many new constraints such as long charging time and battery degradation to \ac{IoEV} compared to traditional \ac{IoV}. Next to the battery layer is the \ac{EV} layer, where \ac{EVs} are considered individually for various aspects such as \ac{EV} charging behaviors and load, as well as an \ac{EV} routing problem. Then, we consider the existence of many \ac{EVs} that collectively share a few charging stations connected to the smart grid, which becomes the third layer. This layer features the aggregated demand and aims to achieve the optimal charging scheduling of many \ac{EVs}. Finally, we provide an investigation of the security layer that is located vertically across the above three layers. For reader's convenience, we also present a list of common abbreviations for reference in \cref{tab:list_of_abbr}. In summary, we make the following main contributions in this paper. 

\begin{itemize}
    \item We present a detailed survey of the latest \ac{GenAI} techniques across all layers of \ac{IoEV}, including an in-depth exploration of various \ac{GenAI} models. We also highlight their roles in solving various IoEV problems such as data scarcity charging load prediction.
    \item We systematically categorize GenAI-enabled \ac{IoEV} applications into four distinct layers for battery, individual \ac{EV}, the grid, and security. We describe each layer with the specific \ac{GenAI} techniques for their respective \ac{IoEV} applications, and provide a holistic view of how \ac{GenAI} can be integrated within the \ac{IoEV} ecosystem.
    \item We provide a summary of publicly available datasets for training \ac{GenAI} models within the \ac{IoEV} context. It facilitates future research on \ac{GenAI} for \ac{IoEV} applications.
    \item We identify and discuss the critical challenges and gaps in the existing \ac{GenAI} applications in \ac{IoEV}. After that, we suggestion future research directions aiming at solving existing challenges and providing new opportunities.
    \item  We bridge the gap between multiple disciplines including: computer science, electrical engineering, and transportation. This comprehensive survey could serve as a valuable resource for a wide audience. 
\end{itemize}

The subsequent sections of this paper are organized as follows. 
In \cref{sect2:Basics}, the basic concepts of \ac{GenAI} and \ac{EV} charging system are initially conducted to give a brief background. 
Following this, a comprehensive review of \ac{GenAI} techniques as they pertain to \ac{IoEV} is undertaken in \cref{sect3:GAI_IoEV}. 
Subsequently, the available public datasets are summarised in \cref{sect4:GAI_IoEV_Dataset}. 
Next, the future directions are recommended in \cref{sect5:Future_Directions}. 
Lastly, a conclusion is presented in \cref{sect6:Conclusions}.

\section{Background}
\label{sect2:Basics}
In this section, we introduce the concept of EV and \ac{IoEV}, the basics of \ac{GenAI} models, as well as a brief introduction of GenAI's industry adoption.

\subsection{Electric Vehicle (EV) and Internet of EV (IoEV)}

\begin{figure}[t!]
    \centerline{
            \includegraphics[width=1.0\columnwidth]{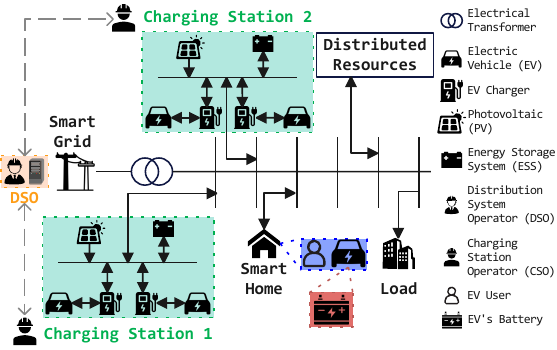}
    }
    \caption{Concept of \ac{EV} charging system in an electrical distribution network: the network consists of charging stations, smart home, distributed resources, and load. \Acf{DSO} and \acf{CSOs} manage the distribution network and the charging stations respectively. Smart homes and charging stations with \ac{PV} and \ac{ESS} can coordinate with \ac{DSO} to ensure a stable and robust smart grid operation. }
    \label{Fig:EV_Charging_System}
\end{figure}

\Cref{Fig:EV_Charging_System} shows the concept of an \ac{EV} charging system in an electrical distribution network. The network consists of charging stations, residential buildings (e.g., smart homes), load, and distributed resources (e.g., renewable energies, and grid-scale energy storage systems). A \ac{DSO} being one of the grid operators, manages the electrical distribution network. The \ac{DSO} can coordinate with \ac{CSOs} and smart home owners to optimally schedule the charging/discharging of connected \ac{EVs}. In this case, the \ac{DSO} sends requests to the contracted \ac{CSOs} and smart home owners to increase/decrease the energy consumption in a certain period of time. Then, the \ac{CSOs} and smart home owners respond to the \ac{DSO}'s request and optimally schedule the \ac{EV} charging/discharging, considering \ac{PV} generation and an \ac{ESS} for achieving different objectives such as charging cost minimization. To participate in the day-ahead energy market, \ac{CSOs} often need to forecast the electricity price and \ac{EV} load for optimal \ac{EV} scheduling in advance. Moreover, \ac{EV} charging behaviors such as arrival time, departure time, charging duration, and unplugging time can influence \ac{EV} load forecasting. From the \ac{EV} users' perspective, they may not only be interested in saving charging costs through smart home systems but also need to understand \ac{EV}'s battery status. Two important aspects are \acf{SoC} and \acf{SoH}. The former refers to the remaining quantity of electricity available in the \ac{EV} battery, which implies the remaining range. The latter indicates the aging status of the battery for making decisions about battery maintenance and retirement \cite{web_SOC_SOH_defination}. Furthermore, an \ac{IoEV} ecosystem, is formed by connecting \ac{DSO}, \ac{CSOs}, and \ac{EV} users, and the optimal \ac{EV} scheduling can be achieved through coordinated efforts among the \ac{IoEV} components. The approach ensures that the concerns of a \ac{DSO}, \ac{CSOs}, and \ac{EV} users are well considered and addressed based on their respective interests and constraints.

\subsection{Basics of Generative Artificial Intelligence (GenAI)}

\Ac{GenAI} is proposed based on traditional \ac{ML}, which are discriminative models that learn the probability distribution $p(y|x)$ in Bayes' theorem for input $x$ and output $y$. The discriminative models categorize the data space into different classes by learning the decision boundaries. They often focus on distinguishing between different classes or outcomes. In the context of computer vision applications, the discriminative models are incapable of processing unknown inputs and it is required to provide label distributions for every image. As such, traditional \ac{ML} models are mainly used for classification, regression, clustering, etc.

\acs{GenAIs} are different from traditional \ac{ML} and they learn the probability distribution of data $p(x)$ for unconditional generative models or $p(x|y)$ for conditional generative models. This allows \acs{GenAIs} to understand the underlying data distribution and generate new samples from the distribution. The generated data/contents could be statistically similar to the input data and the similarity is useful for data augmentation, simulation, and creative tasks. Overall, the generative nature of \ac{GenAI} is useful for developing more dynamic and innovative solutions across various domains.

\begin{figure*}[ht!]
    \centerline{
            \includegraphics[width=1.0\textwidth]{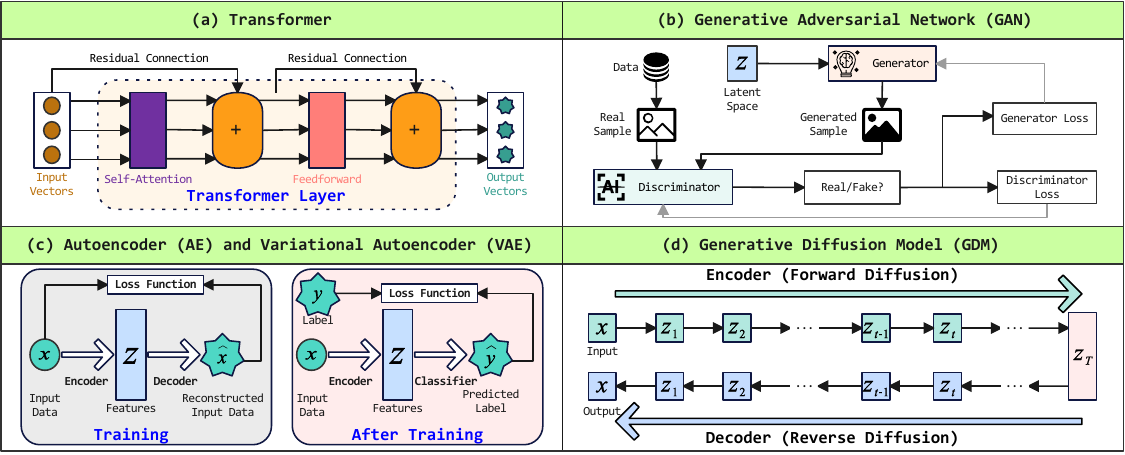}
    }
    \caption{Concept of basic \ac{GenAI} models \textemdash \, Transformer, \ac{GAN}, \ac{AE}, \ac{VAE}, and \ac{GDM}: 
    (a) shows a single-layer Transformer process where the output vectors are achieved by passing the input vectors through self-attention and feedforward layers with residual connections; 
    (b) illustrates principles of \ac{GAN} where the generator competes with the discriminator by producing increasingly realistic samples to \enquote{fool} the discriminator, while the discriminator attempts to differentiate between real and fake data; 
    (c) depicts principles of \ac{AE}: the left side shows the training process of \ac{AE}, while the right side depicts its usage once the model is completely trained; \Ac{VAE} is a subcategory of \ac{AE}, but it differs slightly in that \ac{VAE} uses the mean and diagonal covariance to generate samples in both the encoder and decoder; 
    (d) displays the processes of \ac{GDM} consisting of forward diffusion and reverse diffusion.
    }
    \label{Fig:GenAI_all_in_1}
\end{figure*}

\subsubsection{Transformer}

The Transformer architecture was first introduced by the influential article \enquote{Attention is all you need} \cite{vaswani2017attention_R31r18} which utilizes a self-attention mechanism to capture long-range dependencies without relying on sequential processing. \Cref{Fig:GenAI_all_in_1} (a) shows a single-layer Transformer that includes a typical self-attention module and feedforward layers with residual connections where each layer begins with the application of self-attention. The resulting output from the attention mechanism is then processed by feedforward layers, where the same feedforward weight matrices are used independently for each position. After processing through the first feedforward layer, a nonlinear activation function, e.g., \texttt{ReLU}, is applied. 
 
The Transformer structure is widely used in \ac{LLMs} \cite{LiuFang1}, image processing \cite{zhengyi1}, automatic speech recognition \cite{radford2022robustspeechrecognitionlargescale}, visual question answering \cite{kim2021viltvisionandlanguagetransformerconvolution}, sentiment analysis \cite{Ref25}, etc. Besides the traditional \ac{ML} applications such as audio, computer vision, and \ac{NLP} applications, Transformer is also extended to other domains such as the anomaly detection of the \ac{EV} battery \cite{Ref41}, \ac{EV} routing \cite{Ref24}, and \ac{EV} charging load forecasting \cite{Ref31}. The unique strength of capturing long-range dependencies makes Transformer suitable for parallel computing, scalable according to the different tasks, and transferable with a pretrained model. However, the Transformer is not perfect. For example, the Transformer used in \ac{LLMs} often requires significant computational and memory resources during the training. The performance of the Transformer may be compromised when there is only a small amount of dataset available for training.

\subsubsection{Generative Adversarial Network (GAN)}
The \ac{GAN} model \cite{goodfellow2014generative} consists of two \ac{DNN}: a generator and a discriminator. Both networks engage in an adversarial training process; one network generates new data while the other assesses whether the data is real or fake. \Cref{Fig:GenAI_all_in_1} (b) shows the principle of \ac{GAN}. Assume that the data $x_{i}$ is extracted from the data distribution $p_{data}(x)$, with the goal of sampling it according to $p_{data}$. The sample $z$, the latent variable, drawn from the simple prior $p(z)$ is fed into the generator network. Then, the generated sample is drawn from the data distribution of the generator $p_{g}$. Subsequently, the joint training of the generator and discriminator is carried out until $p_{g}$ converges to $p_{data}$, i.e., $p_{g}\approx p_{data}$. In the training process, the generator network is trained to \enquote{deceive} the discriminator that concurrently learns to classify whether the generated data is real or fake with generator and discriminator loss functions. From the mathematical point of view, it is similar to a two-player minimax game with an objective function. 

\Ac{GAN} was first proposed in 2014 \cite{goodfellow2014generative}. Some evolved versions are \ac{DCGAN} \cite{radford2016unsupervised} in 2016, Progressive \ac{GAN} \cite{karras2017progressive} and Wasserstein \ac{GAN} \cite{arjovsky2017wassersteingan} in 2017, StyleGAN \cite{karras2019style} in 2019 as well as its adoption in semantics communication \cite{Semantic1,Semantic2}, and MaskGAN \cite{lee2020maskgan} in 2020. Besides its applications in computer vision, \ac{GAN} is also extended to other domains, e.g., data augmentation for battery's \ac{SoC} estimation \cite{Ref32}, data augmentation for \ac{EV} charging behaviors \cite{Ref27}, data augmentation for \ac{EV} charging load profile \cite{Ref19}, \ac{EV} load forecasting or changing scenarios generation \cite{Ref33}, as well as generation and detection of adversarial attacks \cite{Ref36}. Moreover, the idea of \ac{GAN} combined with imitation learning formed a new term called \ac{GAIL} \cite{Ref29} which is particularly useful in \ac{RL} when defining a reward function is explicitly difficult. Besides, a variant of GAN is \ac{GAIN} \cite{yoon18a_GAIN}, which is particularly useful for missing data imputation and accordingly improves performance. This is important for IoEV applications that involve incomplete datasets, e.g., with missing entries in charging behavior logs or irregularities in electricity consumption records. While \Ac{GAN} is capable of learning complex and high-dimensional data distribution as well as generating high-quality data, there are challenges, such as balancing generator and discriminator, avoiding mode collapse, and accelerating convergence.  

\subsubsection{Autoencoder (AE) and Variational Autoencoder (VAE)}

An \Ac{AE} is an unsupervised approach that extracts feature vectors from raw data $x$ without labeled examples. It consists of the encoder and decoder during the training as shown in the left side of \cref{Fig:GenAI_all_in_1} (c). The encoder learns the useful information, i.e., features $Z$, from the input raw data $x$. The encoder could be \texttt{Sigmoid}, fully connected, or \texttt{ReLU} \ac{CNN}. After that, the decoder utilizes the learned features to reconstruct the input data $\hat{x}$. The decoder could be \texttt{Sigmoid}, fully connected, or \texttt{ReLU} \ac{CNN} (up-convolution or transposed convolution). The loss function of the training could be the L2 distance between input and reconstructed data. After the training, the decoder is removed and the trained encoder is useful for the downstream tasks as shown on the right side of \cref{Fig:GenAI_all_in_1} (c). For example, a supervised classification model can be initialized using the encoder which is often fine-tuned jointly with the classifier and a task-specific loss function. 

\Ac{AE} can be used as the context encoder in semantic inpainting tasks \cite{pathak2016context}, the temporal context encoder for video applications \cite{liu2017video}, and representation learning \cite{chen2020big}. Besides, \ac{AE} is useful in \ac{EV} related applications, e.g., detection of \ac{FDIAs} that may pose a threat to the \ac{EV} charging process \cite{Ref34}; cyber and physical anomaly detection for abnormal behaviors within \ac{EV} charging stations \cite{Ref44}; detection of \ac{DoS}, fuzzy, and impersonation attacks to the \ac{CAN} protocol communications of \ac{EVs} connected to \ac{EV} charging system \cite{Ref47}. \Ac{AE} offers several benefits. It can effectively reduce the dimensionality of the data and learn the compact representations, which enables it to be useful for data preprocessing and feature extraction. It allows unsupervised learning since no labeled data is required for training, and is suitable for identifying anomalies by learning to reconstruct normal data well. However, vanilla \ac{AE} lacks generative capabilities compared to \ac{VAE} and \ac{GAN}, and cannot capture complex data distributions effectively.

The \ac{VAE} is a type of directed model that relies on approximate inference learned during training and can be optimized solely through gradient-based methods. The concept of \ac{VAE} is similar to \ac{AE}, as it represents a specific subset of \ac{AE}. The encoder and decoder of the original \ac{AE} are modified in the \ac{VAE} where the sampling processes are achieved using means and diagonal covariances. The \ac{VAE} has been used for common \ac{ML} applications such as facial expression editing \cite{yeh2016semantic}, future forecasting from static images \cite{walker2016uncertain}, and point cloud completion \cite{pan2021variational}. Its usage has also been extended to \ac{IoEV} applications, e.g., anomaly detection for EV's battery \cite{Ref20} and data augmentation for \ac{EV} load profiles \cite{Ref21}. Generally, \Ac{VAE} has better generative capabilities compared to vanilla \ac{AE}, and can generate new samples similar to the training dataset. The capabilities make \ac{VAE} suitable for data augmentation and image synthesis, and useful for the clustering and interpolation tasks. However, using a Gaussian prior and a reconstruction loss function, \ac{VAE} may not be able to capture fine details well, e.g., in images with blurry outputs. Moreover, \ac{VAE} may suffer from mode collapse where the model generates a few types of outputs despite having diverse data. Its generative capabilities can be also limited by the latent space Gaussian assumption. 

\subsubsection{Generative Diffusion Model (GDM)}
\Cref{Fig:GenAI_all_in_1} (d) shows the principle of the \ac{GDM}. In its forward diffusion process, i.e., encoder, the model transforms $x$ through a sequence of latent variables $z_1$, \ldots, $z_T$. The procedure is predefined and progressively blends the data with noise until only noise persists at $z_T$. Given a sufficient number of steps, both the conditional distribution $q(z_T|x)$ and the marginal distribution $q(z_T)$ of the final latent variable converge to the standard normal distribution. All the learned parameters are included in the decoder as predefined. In the reverse diffusion process, the data is processed through the latent variables by the decoder which is trained to eliminate noise progressively at each stage. The backward mapping between each pair of adjacent latent variables $z_t$ and $z_{t-1}$ is achieved through the training of a sequence of networks. Each network is guided by the loss function to perform an inversion of its associated encoder step. Following the training process, the new examples are created by sampling vectors of noise $z_T$ and then processing these through the decoder. 

The diffusion model has been widely utilized in image applications, e.g., the \ac{DDPM} \cite{ho2020denoising} for generating high-quality image samples without adversarial training, \ac{DDIM} \cite{song2020denoising} for improving the sampling speed of \ac{DDPM}, stable diffusion \cite{rombach2022high} for generating images from text, and ControlNet \cite{zhang2023adding} allowing model being trained with a small dataset of image pairs. Furthermore, the diffusion model has been extended for other applications, e.g., network optimization \cite{Ref_m11_Du}, generating optimal pricing strategies \cite{Jiacheng1}, repairing and enhancing extracted signal features in wireless sensing \cite{Jiacheng2}, estimating the signal direction of arrival in near-field scenarios \cite{Jiacheng4}, estimating battery's \ac{SoH} \cite{Ref37}, and generating \ac{EV} charging scenarios \cite{Ref26}. \Ac{GDM} can produce high-resolution and realistic samples, achieving similar performance of \ac{GAN} generated data or even better, e.g., outperforming the traditional \ac{GMM} model by 91\% for charging scenarios generation. Compared to \ac{GAN}, \ac{GDM} shows better training stability in versatile applications, e.g., optimization \cite{Ref_m11_Du,Jiacheng1}. Moreover, the clear and iterative process of refining the generated data from noise to coherent output makes the generation process of \ac{GDM} more interpretable. Nevertheless, the iterative nature of the process can be the drawback of \ac{GDM}. It requires a significant amount of computational resources for the training and inference of \ac{GDM}. This nature also slows down the sampling process of \ac{GDM} compared to the \ac{VAE} and \ac{GAN} models that can generate samples in a single pass. When designing and tuning the \ac{GDM}, the noise schedules and model architecture have to be carefully considered. This increases the complexity of training well-performed models. Moreover, \ac{GDM} lacks controllability in specific attributes or features. 

\subsubsection{GenAI Development and Deployment}
Three important stages are training, fine-tuning, and deploying the above-mentioned GenAI models in practice. First, GenAI models are trained on large and diverse datasets to learn broad patterns and general features. Foundation GenAI models are produced in this stage and serve as the backbone of various applications. However, such models are general-purposed, so a fine-tuning stage is needed to customize the models for specific applications, such as EV charging, load forecasting, and route optimization. With the support of the backbone and given the realistic constraints, the stage involves small-scale datasets only which, however, shall be domain-specific. Finally, the fine-tuned models are deployed in practice for real-time inference which is significantly less computing intensive compared to training and fine-tuning models. Overall, the three stages, with different resource demands and objectives, orchestrate GenAI development and deployment. 

\subsubsection{Summary}
Transformer, \ac{GAN}, \ac{AE}, \ac{VAE}, and \ac{GDM} are foundational models of \ac{GenAI} which have demonstrated their versatility across a wide range of applications. They were first invented for traditional \ac{ML} tasks such as computer vision and \ac{NLP} and gradually adapted for electric mobility applications. For EV batteries, GenAI helps estimate battery capacity and health and detect potential anomalies. For EVs, GenAI can be used to understand charging behavior, energy management, and routing. With many EVs forming an IoEV, data augmentation and large-scale charging scheduling can be supported by GenAI. Furthermore, GenAI's role in detecting cyber and physical attacks on IoEV components can be explored. Overall, adapting foundational GenAI models for various \ac{IoEV} aspects from batteries to security requires careful consideration of GenAI's respective strengths and limitations. 

\subsection{\Ac{GenAI} for Industry}
With the basics of GenAI, we present adoption of AI and GenAI in EV industry as well as other domains. 

\subsubsection{EV Industry}
GenAI has not yet been specifically presented in the EV industry but AI has been among the strategic focuses of the big EV players and we introduce some latest progress from two major players. AI has been adopted in various aspects of Tesla's business, e.g., EV manufacturing and autonomous driving. The large-scale AI adoption is enabled by Tesla's Dojo supercomputer which offers abundant computing resources and realizes computational-intensive tasks such as high-throughput EV video processing. The company also owns an overarching aspiration to develop \ac{AGI}. BYD has introduced its XUANJI Architecture, an intelligent vehicular framework, integrating electrification with intelligent functionalities and functioning as the EV's cognitive core. EV's internal and external environments are monitored in real-time and the collected information is used to make decisions about the EV's operation to improve safety and comfort. The industry favors the integration of AI and physical systems and we foresee an increasing adoption of GenAI in the EV industry.

\subsubsection{Other Industry Sectors}
Besides EV and the transportation sector, GenAI has found widespread adoption in other industry sectors and we introduce a few sectors as follows. For business and finance, a survey \cite{GenAI_finance} is available with a description of GenAI's practical applications and cutting-edge tools in the sector. The survey \cite{GenAI_healthcare} is about GenAI's applications, advantages, and obstacles for the healthcare sector. The education sector is witnessing GenAI's significant impact and the authors in \cite{GenAI_education} discuss GenAI's ability to boost learner engagement and motivation, emphasizing the need for ethical guidelines and human oversight as well as GenAI's impact on critical thinking. Compared to different industry sectors, the IoEV industry has various unique features, e.g., battery and grid integration. The features require GenAI algorithms to be customized and specialized for performance maximization.

\section{Technical Reviews: Generative Artificial Intelligence (GenAI) for Internet of Electric Vehicles (IoEV)}
\label{sect3:GAI_IoEV}
In this section, we provide an overview and discussions of \ac{GenAI}'s application in different layers of \ac{IoEV}.  

\subsection{Layer 1: Electric Vehicle (EV)'s Battery}
The battery is the core component of an \ac{EV}. It powers an electric motor and has a direct impact on EV's range, performance, and efficiency. It is also the most expensive part of most \ac{EVs}, due to the fact of which, battery's operating condition and longevity play a crucial role in the overall user experience and sustainability of the \ac{EVs}. We specifically would like to survey three important aspects of batteries, including anomaly detection \cite{Ref20, Ref41}, \ac{SoC} \cite{Ref30, Ref32, Ref42}, and \ac{SoH} \cite{Ref37}. For anomaly detection, various detection algorithms can be integrated with the \ac{BMS} for proactive battery maintenance before any potential failures cause hazardous battery damage. The \ac{SoC} and \ac{SoH} indicate the battery's short-term energy capacity and long-term health condition, respectively. Specifically, \ac{SoC} measures the stored energy relative to the maximum capacity and \ac{SoH} reflects the battery's maximum capacity which degrades over time. Both aspects are influenced by different factors such as temperature, charging voltage, and cycling history \cite{web_SOC_SOH_defination}.

\subsubsection{Anomaly Detection}
Despite continuous technological progress in the past years, battery safety remains a big concern for EV owners and customers. One of the most hazardous issues is battery fire, the occurrence of which has raised debates and doubts about battery safety and highlights the necessity of early anomaly detection to prevent potential safety breaches and irreversible damage \cite{web_battery_fire}. 

\paragraph{Traditional Machine Learning (ML) Approaches}
The early effort of battery anomaly detection and diagnosis is based on traditional \ac{ML} algorithms, e.g., the random forest \cite{naha2020internal}, multiclass relevance vector machine \cite{xie2020quantitative}, and finite-element-based models \cite{jia2021_R41r40}. These \ac{ML} algorithms are generally simple to implement but the performance suffers when the raw data is used directly. Domain knowledge complements the capability of the algorithms by guiding the extraction of domain-specific and useful features as the \ac{ML} input, with improved correlation with battery anomalies. The advancements of deep learning offer new methods for battery anomaly detection. Among the methods, \ac{LSTM} should be the most widely used architecture \cite{hong2019fault_R41r43,hong2019R41r44}, which is capable of predicting battery voltage with multiple inputs \cite{hong2019fault_R41r43} and forecasting parameters such as voltage, temperature, and \ac{SoC} simultaneously \cite{hong2019R41r44}. However, \ac{LSTM} being a type of \ac{RNNs} presented challenges in practical training scenarios, with poor training stability and issues such as vanishing or exploding gradients. 

\paragraph{AE and VAE-based Approaches}      
\Ac{GenAI} can potentially address the above-mentioned challenges and the reconstruction-based models have been studied. \Ac{AE} as a basic reconstruction model is shown to be ineffective for generating diverse and high-quality data. This is largely due to the deterministic nature of the latent codes produced by the \ac{AE} encoder. A more suitable model is \ac{VAE}, which adeptly learns the probability distribution of \ac{MVTS} to be robust against perturbations and noise \cite{MVTS_Zhao_2020}. In \cite{Ref20}, a semi-supervised VAE-based anomaly detection model was proposed for early anomaly detection in battery packs. The model detects different anomalies such as irregular terminal voltage, differences between all bricks, and abnormal temperature fluctuations. The model, named \acs{GRU-VAE}, consists of a \ac{GRU} and a \ac{VAE}, where the former captures \ac{MVTS} and the latter reconstructs the input samples. Worth mentioning that the paper is based on an \ac{EV} operation dataset from the \ac{NSMC-EV} in Beijing \cite{li2020battery}. The dataset includes 13-dimensional time series signals such as data acquisition time, vehicle speed and state, charging state, voltage, current, mileage accumulated, \ac{SoC}, temperature, insulation resistance, and DC–DC state. \acs{GRU-VAE} shows an improvement in anomaly detection compared to the AE-based models, achieving a 25\% increase in F1-score \cite{Ref20}. Nevertheless, the model may need to be updated, given the changed data patterns over time. 

\paragraph{Transformer-based Approach} 
Anomaly detection can also be transformer-based \cite{Ref41}. The authors in \cite{Ref41} developed \acs{BERTtery}, a transformer-based model for battery fault diagnosis and failure prognosis. \Acs{BERTtery} is able to capture early-warning signals across multiple spatial–temporal scales in various operational conditions, predict battery system fault evolution using onboard sensor data, and avoid faults leading to thermal runaways \cite{web_ThermalRunaway}. The model is based on a dataset with various battery faults and failures, e.g., internal short circuits, lithium plating faults, overcharging/overdischarging, abnormal self-discharge, abnormal capacity degradation, abnormal voltage fluctuations, abnormal temperature behaviors, electrolyte leakages, cell balancing issues, and thermal runaways \cite{zhao2022data_R41r41}. Specifically, the model's input includes the time series of voltage, current, and temperature, sampled every 10 seconds from real-world \ac{EV} operations. The output of the model is the predicted safety labels. For future work, the generalization ability of the proposed model could be improved by considering different battery types and operational conditions, to enhance the early warning capabilities with reduced false alarms.

\subsubsection{State of Charge (SoC) Estimation}
SoC indicates how much battery energy remains and directly affects the EV's range \cite{web_SOC}. Thus EV owners monitor SoC for trip planning, charging scheduling, and battery usage optimization.

\paragraph{Traditional ML-based Approaches}
Many \ac{ML} algorithms have been adopted for \ac{SoC} modeling, e.g., the random forest \cite{MAWONOU2021229154}, \ac{GPR} \cite{WANG2023232737}, and \ac{SVM} \cite{ZHAO2022124468}. These algorithms are commonly adopted partially because of their simplicity, which however is one of the reasons that they cannot handle the complex battery operating conditions. For relatively more complex algorithms, the \acs{CNNs} \cite{FAN2022124612} capture local feature representation for time series forecasting and may overlook distant variable correlations, limiting the ability to capture remote topological structures. The long dependencies can be learned by \ac{LSTM} \cite{8695733}, which however is limited with extended sequences due to the inherent constraint of recurrent models, where signals must traverse both forward and backward. Moreover, traditional \ac{ML} models typically require extensive data for model training; otherwise, the model could be over-fitted with reduced accuracy. Unfortunately, data is often scarce. One idea is to use \ac{GenAI} for data augmentation, and this idea has been studied for \ac{SoC} estimation, as detailed below.  

\paragraph{GAN-based Approaches}
Efforts to utilize \ac{GenAI} for general-purpose data augmentation have yielded successes, e.g., \ac{CR-GAN} \cite{CR_GAN2021}, \ac{RC-GAN} \cite{RC_GAN2022}, 
\ac{TimeGAN} \cite{TimeGAN_Naaz2021}, and global trend diffusion algorithms \cite{Diffusion_Yong2023}. Among the \ac{GAN} variants, \Ac{CR-GAN} and \ac{RC-GAN} struggle to capture the temporal dynamics encompassed within the entirety of a time series sometimes, owing to gradient vanishing or explosion that arose when processing long sequences with \ac{RNNs}. \Ac{TimeGAN} divides long time series into smaller segments, potentially leading to the loss of crucial cross-segment information. The global trend diffusion algorithm lacks sufficient adaptability to different scenarios because of its triangle distribution and fails to adequately represent the complexity of the original data. 

Several GAN-based models are specially designed for \ac{SoC} estimation. The \ac{WGAN} is a promising base model, which is shown to be robust by generating the underlying real data distribution, enhancing the generation quality of vanilla \ac{GAN}, and accelerating convergence \cite{8629907}. Hence, the authors in \cite{Ref30} developed a \ac{TS-WGAN} based on \ac{WGAN} for \ac{SoC} estimation. The new model consists of data pre-processing and a \ac{WGAN-GP} architecture \cite{8629907}. Besides, the model is trained by the \ac{EV} dataset \cite{en14123692} and LG 18650HG2 Li-ion Battery dataset \cite{8790543}. The former includes \ac{EV} information during charging/discharging such as timestamp, vehicle speed, voltage, current, cell temperature, motor controller input voltage and current, mileage, and \ac{SoC}. The latter consists of the battery’s performance data during charging, discharge cycle measurements, and drive cycles. Note that \ac{TS-WGAN} may suffer from convergence issues as it requires complex index modifications due to the nature of GANs.

Another \ac{WGAN}-based model is proposed in \cite{Ref42}, named \ac{C-LSTM-WGAN-GP}. It is an LSTM-based conditional \ac{GAN} model and owns the capability to generate data that closely resembles actual battery data across various profiles. The training data is obtained from the experiments. Two Li-ion rechargeable cells were set up for the experiments where one is Li-ion with the material of Ni/Co/Mn ternary composites and another one is Li-ion with the material of lithium iron phosphate. The usable battery information, such as terminal voltage, current, and temperature, was monitored and recorded while \ac{SoC} was estimated after the experiment was done. Future improvements can be architecture and training convergence optimization, as well as system integration with existing BMS.

Extending from the above \ac{WGAN}-based models, the authors in \cite{Ref32} introduced a \ac{TS-DCGAN} framework. The framework combines the time-frequency domain techniques and \ac{DCGAN} \cite{radford2016unsupervised} to train the models for \ac{SoC} estimation. The models generate synthetic datasets with high fidelity and diversity, effectively capturing the dependencies between multidimensional time series. The training data is obtained from LG INR18650HG2 batteries \cite{web_TOMObattery}, which are common \ac{EV} batteries \cite{Ref32}. The dataset consists of battery voltage, current, cell temperature, \ac{SoC} at different temperatures, and driving cycles. Real data is complemented with the generated synthetic data. The experimental results show that \ac{TS-DCGAN} successfully reduces the discriminative score of CR-GAN by 53\% and TimeGAN by 46\% in producing reliable synthetic datasets for the subsequent \ac{SoC} estimation tasks. Generally, the common issues of the GAN-based models, such as training stability, also apply to \ac{TS-DCGAN}, which can be further enhanced.

\subsubsection{State of Health (SoH) Estimation}
\ac{EV} owners are not only interested in knowing the battery's \ac{SoC} but also \ac{SoH} \cite{web_SOC_SOH}. This allows them to plan effectively and schedule maintenance or replacement as needed. It also offers vital insights into \ac{EV} battery control strategies, protection mechanisms, and sustainable development \cite{Bat_SOH2021}. The \ac{SoH} estimation methods can be categorized into model-driven and data-driven approaches. The former includes the electrochemical model \cite{LI2020115104} and the equivalent circuit model \cite{Amir2022_circuit_SOH}. Relatively, the latter has advantages such as independence from prior knowledge of battery mechanisms and avoidance of subjective intervention, with a focus on latent input-output relationships.

\paragraph{Traditional ML Approaches}
SoH prediction has been addressed by different traditional \ac{ML} algorithms, e.g., \ac{ANN} \cite{ANN_SOH2013}, \ac{SVM} \cite{DENG2016_SVM}, \ac{AutoML} \cite{Luo2022_AutoML}, \ac{GPR} \cite{LI2019_GPR}, \ac{CNN} \cite{SOHN2022_R37r18}, \acs{RNN} \cite{harada2020rnn_R37r19}, \ac{LSTM} \cite{LSTM2022_R37r20}, \ac{GRU} \cite{GRU_R37r21}, and \ac{BNN} \cite{yunyi1}. The above models are regarded as discriminative models, which focus on battery health indices including cycles and capacity. They map input parameters to output variables without prior sample knowledge, and adjust network weights through training with a loss function. However, it is challenging for them to well capture intrinsic characteristics to represent the battery's operational dynamics accurately, and the challenge is often addressed by increasing the quality and quantity of the training dataset.

\paragraph{Diffusion-based Approach}
Compared to conventional discriminative \ac{ML} models, \acs{GDMs}, can capture the distribution characteristics inherent in training data more accurately, thereby offering a more comprehensive understanding of the underlying problem. By leveraging generative diffusion techniques, one could mitigate the risk of introducing significant deviations to the overall distribution of training data, thereby facilitating robust modeling that transcends merely isolated feature representations \cite{yang2023diffusion}. 

The authors in \cite{Ref37} introduced a diffusion-based model namely, \ac{DDPM}, to predict the \ac{SoH} of lithium-ion batteries with both offline and online modeling. The dataset used in their experiment is a lithium iron phosphate battery dataset from TOYOTA Research Institute \cite{RN713} and a nickel cobalt manganese battery dataset from their laboratory \cite{Ref37}. The datasets consist of specifications such as rated capacity, number of cells, charging/discharging current, maximum/minimum cut-off voltage, and number of cycles. The prediction variables are the battery's capacity in the unit of Ah. The proposed \ac{DDPM} outperforms other \ac{ML} techniques in terms of several \ac{SoH} prediction error metrics, e.g., \ac{RMSE}, \ac{MAE}, and \ac{MAPE}. Taking \ac{RMSE} as an example, \ac{DDPM} can reduce prediction errors of \acs{RNN} by 57\%, \ac{LSTM} by 35\%, \ac{GRU} by 10\%, transformer \cite{Transformer2022_SOH} by 70\%, and CNN-Transformer \cite{li2019enhancing_R31r17} by 52\% \cite{Ref37}. In the future, \ac{DDPM}'s effectiveness for SoH may further be explored by comparing it with other \ac{GenAI} methods such as \ac{VAE} and \ac{GAN}.

Overall for layer 1, we provide a summary of the \ac{GenAI} works in \Cref{table:compare_layer1}. From the table, we can see that the VAE-based \cite{Ref20} and transformer-based \cite{Ref41} models can be utilized for battery anomaly detection. Moreover, GAN-based models, e.g., \ac{TS-DCGAN} \cite{Ref32} and \ac{C-LSTM-WGAN-GP} \cite{Ref42}, are developed mainly for data augmentation to enhance the accuracy of \ac{SoC} estimation. And \ac{TS-WGAN} \cite{Ref30} considers both data augmentation and \ac{SoC} estimation. Furthermore, the recent advancements in generative diffusion models have been applied in \ac{DDPM} \cite{Ref37} for \ac{SoH} estimation.

\begin{table*} 
\caption{Summary of \ac{GenAI} models for \ac{IoEV} in layer 1 for batteries. \\ \textcolor{cyan}{\dmark}: \ac{GenAI} methods; \textcolor{green}{\cmark}: pros of the methods; \textcolor{red}{\xmark}: cons of the methods.}
\label{table:compare_layer1} 
\begin{tblr}{
  width = \linewidth,
  colspec = {m{0.09\linewidth}m{0.07\linewidth}m{0.1\linewidth}m{0.635\linewidth}}, 
  cells = {c},     
  hlines,
  vline{2-4} = {-}{},
  vline{2} = {2}{-}{},
  hline{1,8} = {-}{0.08em},
}
\hline
\textbf{Application} & 
\textbf{Reference} & 
\textbf{\acs{GenAI} Model} & 
\textbf{Pros \& Cons} \\
\SetCell[r=2]{c} Anomaly Detection &   
\cite{Ref20} & 
\acs{GRU-VAE}   &    
\begin{minipage}[c]{\linewidth}
    \begin{itemize}[noitemsep,topsep=0pt,leftmargin=8pt]
       \item[\textcolor{cyan}{\dmark}] A \acs{GRU-VAE} framework for battery anomaly detection.
       \item[\textcolor{green}{\cmark}] Learn probability distribution of multivariate time series data adeptly.
       \item[\textcolor{green}{\cmark}] Robust against perturbations and noise.
       \item[\textcolor{red}{\xmark}] Adaptability issues and frequent model updates for new data.
    \end{itemize}
\end{minipage} \\
%
Anomaly Detection & 
\cite{Ref41} & 
\acs{BERTtery} & 
\begin{minipage}[c]{\linewidth}
    \begin{itemize}[noitemsep,topsep=0pt,leftmargin=8pt]
        \item[\textcolor{cyan}{\dmark}] A transformer-based method for battery fault diagnosis and failure prognosis.
        \item[\textcolor{green}{\cmark}] Learn battery's nonlinear cell behaviors in a self-supervised data-driven manner.
        \item[\textcolor{green}{\cmark}] Competitive performance, e.g., above 95\% in accuracy, precision, recall and F1 score.
        \item[\textcolor{red}{\xmark}] Lack generalization ability to various battery types and operational conditions.
        \item[\textcolor{red}{\xmark}]  Need for an improved early warning predictions with reduced false positives and negatives.
    \end{itemize}
\end{minipage} \\

\SetCell[r=3]{c} \acs{SoC} Estimation  &    
\cite{Ref30} & 
\acs{TS-WGAN} & 
\begin{minipage}[c]{\linewidth}
    \begin{itemize}[noitemsep,topsep=0pt,leftmargin=8pt]
        \item[\textcolor{cyan}{\dmark}] A GAN-based approach for \acs{SoC} estimation of lithium-ion batteries.
        \item[\textcolor{green}{\cmark}] Robust and able to generate underlying data distribution.  
        \item[\textcolor{green}{\cmark}] Enhanced generation quality of vanilla \acs{GAN} and accelerated convergence.
        \item[\textcolor{red}{\xmark}] Convergence issue during training and requiring complex index modifications.
        \item[\textcolor{red}{\xmark}] Limitations for real-time \acs{SoC} estimation due to its computational intensity.
    \end{itemize}
\end{minipage} \\

\acs{SoC} Estimation & 
\cite{Ref32} & 
\acs{TS-DCGAN} & 
\begin{minipage}[c]{\linewidth}
    \begin{itemize}[noitemsep,topsep=0pt,leftmargin=8pt]
        \item[\textcolor{cyan}{\dmark}]  A GAN-based approach to generate synthetic data for \acs{SoC} estimation.
        \item[\textcolor{green}{\cmark}] Produce synthetic data of high fidelity and diversity.
        \item[\textcolor{red}{\xmark}] Common issues of \acs{GAN}-based structure, e.g., stability of model training.
    \end{itemize}
\end{minipage} \\
\acs{SoC} Estimation  & 
\cite{Ref42} & 
\acs{C-LSTM-WGAN-GP} & 
\begin{minipage}[c]{\linewidth}
    \begin{itemize}[noitemsep,topsep=0pt,leftmargin=8pt]
        \item[\textcolor{cyan}{\dmark}] A LSTM-based conditional \acs{GAN} model for data generation and \acs{SoC} estimation.
        \item[\textcolor{green}{\cmark}] Stable model training performance.
        \item[\textcolor{green}{\cmark}] Generate realistic battery data cross various profiles.
        \item[\textcolor{red}{\xmark}] Need for an improved model architecture and the speed of convergence.
        \item[\textcolor{red}{\xmark}] Lack of implementation on online \acs{BMS}.
    \end{itemize}
\end{minipage} \\
\acs{SoH} Estimation & 
\cite{Ref37} & 
\acs{DDPM} & 
\begin{minipage}[c]{\linewidth}
    \begin{itemize}[noitemsep,topsep=0pt,leftmargin=8pt]
        \item[\textcolor{cyan}{\dmark}] A diffusion-based model for battery \acs{SoH} estimation.
        \item[\textcolor{green}{\cmark}] Low prediction errors compared with \acs{RNN}, \acs{LSTM}, \acs{GRU}, and transformer-based models.
        \item[\textcolor{red}{\xmark}] Need for a comparison with other \acs{GenAIs}, e.g., \acs{GAN} and \acs{VAE} for the same application.
    \end{itemize}
\end{minipage} \\ \hline
\end{tblr}
\end{table*}

\begin{figure}[t] 
	\centerline{            \includegraphics[width=1.0\columnwidth]{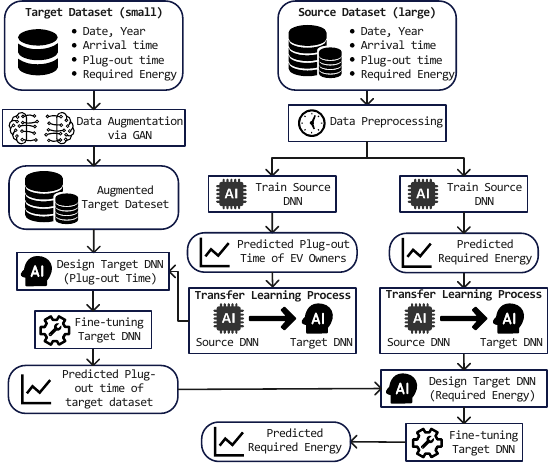}
	}
 	\caption{The flowchart of proposed framework to address the cold-start forecasting problem in predicting the \ac{EV} charging behaviors such as plug-out hour and required energy for newly committed \ac{EVs} \cite{Ref27}.}
    \label{Fig3:GAN_Ref27}
\end{figure}

\subsection{Layer 2: Individual Electric Vehicle (EV)}

\ac{EV} is an integrated system that combines key components like batteries in layer 1. Instead of focusing on one component as the research works for layer 1, the research in the \ac{EV} layer emphasizes the integration and functionality of the \ac{EV} system. \ac{GenAI} has been studied for the \ac{EV} layer for two aspects including charging \cite{Ref27,Ref35} and routing \cite{Ref24}, and we present technical details below. 

\subsubsection{\ac{EV} Charging Behaviors and Loads}
The research on \ac{EV} charging is important for optimizing energy usage, balancing grid demand, and improving charging efficiency. The growing adoption of EVs makes data-driven approaches ideal for related research with a growing amount of EV data generated. The approaches are typically designed for optimizing the charging parameters and analyzing parameter correlations, user travel patterns, and vehicle trajectories \cite{chen2018_R16r4}. 

\paragraph{Traditional ML-based Approaches}
Several \ac{ML} algorithms have been applied in load forecasting and energy management for \ac{EV}s, e.g., \acs{RNN} \cite{jahangir2019_R27r21}, \ac{CNN} \cite{zhang2020_R27r22}, CNN-GRU \cite{jahromi2022_R27r20}, and \ac{LSTM} with \ac{RL} \cite{Ref5}. A common issue of the above algorithms is the demand for extensive training data (to avoid overfitting and underfitting) \cite{forootani2022_R27r24}. Such data demand cannot be satisfied by many \ac{EV} service providers, facing realistic constraints, e.g., only 365 charging samples per year. This is especially true at the beginning stages of data collection, causing the famous \enquote{cold-start forecast problem} \cite{gilanifar2021_R27r25}. Therefore, researchers are looking for methods to generate extensive datasets with relatively small-scale data collected from real EVs, and \ac{GenAI} is a promising method.

\paragraph{GAN-based Approaches}
The above-mentioned cold-start problem has been addressed in \cite{Ref27}. The authors developed a transfer learning-based framework using a deep generative model, \ac{GAN}, to address the problem of predicting \ac{EV} charging behaviors. \Cref{Fig3:GAN_Ref27} shows the flowchart of the proposed framework which integrates \ac{GAN} and \ac{DNN} for the forecasting tasks. As seen from the figure, two source \acs{DNNs} are trained on residential \ac{EV} owner data to predict the plug-out time of \ac{EV} owners and the required energy, respectively. The transfer learning then adapts the knowledge of the tested EV to the target \ac{DNN} (plug-out time) for forecasting new \ac{EV} plug-out hours. The same approach is applied to the target \ac{DNN} for required energy. Also, \ac{GAN} models are used to augment the target dataset. The target \acs{DNNs}, with the weights from the source EV model and the GAN-generated data, are fine-tuned to predict the charging behaviors of target EVs.

Besides, the research is based on a public dataset from EA Technology, which specializes in providing asset management solutions for the owners and operators of electrical assets \cite{web_dataset_Ref27}. The available parameters of residential charging events include dates, arrival hours, plug-out hours, and required energy. Finally, the derived models can achieve significant performance gains over \ac{SVR} \cite{chung2019_R27r18} by 12\%, decision tree regression \cite{shahriar2021_R27r19} by 57\%, \ac{KNNR} by 60\%, \ac{DNN} \cite{jahromi2022_R27r20} by 31\%, and a GAN-DNN based approach \cite{Ref18} by 10\% \cite{Ref27}. For future work, the performance of proposed method could be improved by applying the clustering algorithms for efficient transfer and multi-source datasets for model generalizability.

In \cite{Ref16}, a \ac{CW-GAN} model was developed for generating \ac{EV} charging behavior parameters such as arrival time, departure time, and \ac{SoC}. The study is based on a small dataset of private \ac{EVs} and charging piles in three functional zones (i.e., office, business, and residential areas) \cite{Ref16}. The input of the model includes noise and conditional labels, and the output generates different parameters of \ac{EV} charging behavior. Note that the quality of conditional labels has a significant impact on the overall performance of \ac{CW-GAN}, so accurate label selection is crucial.

Besides charging behavior, \ac{GAN} has also been studied for charging load. In \cite{Ref43}, a GAN-based \ac{HEDGE} tool is introduced to semi-randomly generate synthetic daily profiles of \ac{EV} and household loads as well as \ac{PV} generation. The generated residential energy data spanning multiple days exhibits consistency in terms of magnitude and behavioral clusters. Several profile datasets are used for model training. Household load and solar generation are available in TC1a \cite{web_data_R43r1} and TC5 \cite{web_data_R43r2}, respectively, from the \ac{CLNR} project. The \ac{EV} loads are estimated based on the general population’s travel patterns dataset from the UK's National Travel Survey \cite{web_data_R43r3}. In the future, \ac{HEDGE} will be valuable for subsequent research tasks, e.g., optimal scheduling with HEDGE generated profiles.

\paragraph{Hybrid Approach}
Instead of using \ac{GAN} alone, hybrid solutions have been investigated for synthetic data generation also. In \cite{Ref35}, a \acs{VAE-GAN} model is developed to generate synthetic time-series energy profiles such as EV load profiles in smart homes. The synthetic data is subsequently utilized in the Q-learning-based home \ac{EMS} to maximize long-term profit through optimal load scheduling. The model is trained with the iHomeLab PART dataset \cite{huber2020residential_R35r27}, which includes power consumption profiles of five residences.
The study compared the \acs{VAE-GAN} with a Vanilla \ac{GAN} and a \ac{GMM}, employing the \ac{KL} divergence to assess the distance between real and synthetic data distributions. The results reveal that \acs{VAE-GAN} can achieve 18\% and 33\% performance improvement over \ac{GAN} and \ac{GMM}, respectively, in generating \ac{EV} load data \cite{Ref35}.
The improvement shows that the model is able to learn various smart home data distributions (e.g., electric load, \ac{PV} generation, and \ac{EV} charging load), and generate realistic data samples without prior analysis before training.
Nonetheless, the model relies on the quality and diversity of training data with inconsistent scalability and adaptability. Future research may incorporate diverse datasets to enhance the model's robustness and improve scalability across broader smart grid applications. 

\subsubsection{\ac{EV} Routing}  
The future of urban delivery is likely to be driven by autonomous green vehicles \cite{YINGFEI2022_R24r7}. The trend underscores the significance of efficient route planning, addressed as the \ac{EVRP}. In \ac{EVRP}, an \ac{EV} starts from a designated depot with a partial/full charge to serve customers with time restrictions. Each \ac{EV} can stop at charging stations or return to the depot to recharge. The goal of \ac{EVRP} is to find cost-effective routes for the \ac{EV} fleet subject to battery constraints.

\paragraph{RL-based Approaches}
Recently, researchers have applied supervised learning and \ac{RL} to address the \ac{VRP} amidst the growth of \ac{ML}, e.g., a pointer network \cite{vinyals2015pointer_R24r16}. A significant challenge is to obtain sufficient labels for producing optimal solutions in large-scale problems like \ac{EVRP}. \ac{RL} is a label-free approach so it can be a viable option for addressing large-scale problems \cite{barrett2020exploratory_R24r19}. \Ac{DRL}, as a type of \ac{RL}, has been applied to solve the \ac{VRP}, often utilizing the encoder-decoder architecture in neural network design \cite{pan2023deep_R24r29}, and achieved good performance. However, related research works often focus on basic routing issues, and overlook the complexities of \ac{EVRP} with energy and charging constraints, which are unique compared to traditional \ac{VRP}.

\begin{figure}
	\centerline{
            \includegraphics[width=1.0\columnwidth]{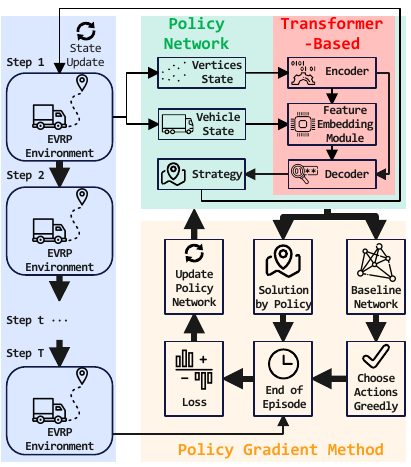}
	}
 	\caption{Framework of \ac{DRL} with Transformer for \ac{EV} routing problem \cite{Ref24}.}
    \label{Fig7:DRL-Transformer_Ref24}
\end{figure}

\paragraph{Transformer-based Approach}
Transformer architecture with an attention mechanism has been proven to be effective in improving computational efficiency and solution quality in solving \ac{VRP} \cite{kool2018attention_R24r22}. Hence, the authors in \cite{Ref24} proposed a Transformer-based \ac{DRL} method for energy minimization of \ac{EVRP}. 
\Cref{Fig7:DRL-Transformer_Ref24} shows the proposed framework where the policy network of \ac{DRL} is modeled by the Transformer's encoder-decoder structure. The features of \ac{EVRP} are captured by the feature embedding module and the policy gradient method is employed for policy training. In the context of an \ac{EVRP} instance, the graph information including the vehicle and the states of vertices undergo encoding by an encoder. This process facilitates the step-by-step construction of \ac{EV} routes by the decoder, leveraging inputs from both the encoder and the feature embedding module. Subsequently, updates to the parameters of the policy network are made based on reward values derived from the policy network and the baseline network.

The \ac{EVRP} instances are created according to the procedure outlined in \cite{ERDOGAN2012100_R24r30}. The locations of customers and charging stations were chosen uniformly at random from a square kilometer area. Finally, the proposed transformer-based \ac{DRL} method is compared with the exact algorithm \cite{web_gurobi}, improved \ac{ACO} \cite{zhang2018electric_R24r12}, \ac{ALNS} algorithm \cite{goeke2015routing_R24r14}, and \ac{AM} \cite{kool2018attention_R24r22} under various scenarios, e.g., different number of \ac{EVs} and charging stations. Specifically for a case study with 100 \ac{EVs}, the proposed method can reduce the energy consumption of \ac{EVs} by 1\% compared to \ac{AM}, and 10\% to \ac{ACO} \cite{Ref24}. Future work includes enhancing the model's efficiency, testing with real-world data, and expanding the model to accommodate more complex scenarios, e.g., multiple vehicle types and different environmental constraints. Other \ac{GenAIs} such as \ac{GAN} and \ac{GDM} can also be considered for navigation and route optimization application \cite{Ruichen1}.

At the end of such EV layer, we would like to mention Autonomous Vehicles (AVs), which often occur with EV together as new concepts of vehicular technology. However, an AV is not necessarily an EV and it can be driven by different power sources including but not limited to electricity. Our focus in this paper is EV and some interesting discussions of utilizing GenAI in AVs are available in \cite{Survey_AutoVehicles1,Survey_AutoVehicles2} about AV-related aspects such as trustworthiness and navigation.

Overall, the \ac{GenAI} applications for \ac{IoEV} in layer 2 are summarized in \cref{table:compare_layer2}. We find that the GAN-based models, GAN-DNN \cite{Ref27} and \ac{CW-GAN} \cite{Ref16}, can be utilized to generate \ac{EV} charging behaviors. For residential applications, the GAN-based models \cite{Ref35,Ref43} are employed for data augmentations, which is important for other applications, e.g., optimal scheduling of an \ac{EV} at home \cite{Ref35}. Researchers also use the Transformer model \cite{Ref24} to solve the \ac{EVRP}, which is different from traditional \ac{VRP} with new \ac{EV} constraints.

\begin{table*} 
\caption{Summary of \ac{GenAI} for \ac{IoEV} in layer 2. \\ \textcolor{cyan}{\dmark}: \ac{GenAI} methods; \textcolor{green}{\cmark}: pros of the methods; \textcolor{red}{\xmark}: cons of the methods.}
\label{table:compare_layer2} 
\begin{tblr}{
  width = \linewidth,
  colspec = {m{0.09\linewidth}m{0.07\linewidth}m{0.1\linewidth}m{0.635\linewidth}}, 
  cells = {c},      
  hlines,
  vline{2-4} = {-}{},
  vline{2} = {2}{-}{},
  hline{1,8} = {-}{0.08em},
}
\hline
\textbf{Applications} & 
\textbf{Reference} & 
\textbf{Techniques} & 
\textbf{Pros \& Cons} \\
%
\SetCell[r=2]{c} EV Charging Behavior Data Augmentation &  
\cite{Ref27}  & 
GAN-DNN &  
\begin{minipage}[c]{\linewidth}
    \begin{itemize}[noitemsep,topsep=0pt,leftmargin=8pt]
        \item[\textcolor{cyan}{\dmark}] A framework to predict \acs{EV} charging behaviors, e.g., plug-out hour and required energy. 
        \item[\textcolor{green}{\cmark}] Address the cold-start forecasting problem when limited training data is available.
        \item[\textcolor{red}{\xmark}] Need clustering algorithms and multi-source datasets for future work.
    \end{itemize}
\end{minipage} \\
%
EV Charging Behavior Data Augmentation & 
\cite{Ref16} & 
\acs{CW-GAN} & 
\begin{minipage}[c]{\linewidth}
    \begin{itemize}[noitemsep,topsep=0pt,leftmargin=8pt]
        \item[\textcolor{cyan}{\dmark}] A GAN-based model to generate charging behavior data for \acs{EVs}. 
        \item[\textcolor{green}{\cmark}] Learn from small samples, expanding dataset while preserving original probability distribution.
        \item[\textcolor{red}{\xmark}] Dependency on quality of conditional labels during the training.
    \end{itemize}
\end{minipage} \\
Smart Home Data Augmentation  & 
\cite{Ref35} & 
\acs{VAE-GAN} with Q-learning  & 
\begin{minipage}[c]{\linewidth}
    \begin{itemize}[noitemsep,topsep=0pt,leftmargin=8pt]
        \item[\textcolor{cyan}{\dmark}] A data generation scheme using a \acs{VAE-GAN} combined with Q-learning-based home \acs{EMS}.  
        \item[\textcolor{green}{\cmark}] Learn various data distributions in a smart home and generate realistic samples.
        \item[\textcolor{red}{\xmark}] Reliance on quality and diversity of training data.
    \end{itemize}
\end{minipage} \\
Residential \acs{EV} Load Generation & 
\cite{Ref43} & 
\acs{GAN} & 
\begin{minipage}[c]{\linewidth}
    \begin{itemize}[noitemsep,topsep=0pt,leftmargin=8pt]
        \item[\textcolor{cyan}{\dmark}] A GAN-based tool for semi-randomly generated data for \acs{EV} load, \acs{PV}, and household demand. 
        \item[\textcolor{green}{\cmark}]  Generate profiles that keep both magnitude of profile and behavioral consistency over time.
        \item[\textcolor{red}{\xmark}] Need other countries' datasets to extend the capability of the \acs{HEDGE} tool.
    \end{itemize}
\end{minipage} \\
%
\acs{EVRP} & 
\cite{Ref24} & 
Transformer-based \acs{DRL} & 
\begin{minipage}[c]{\linewidth}
    \begin{itemize}[noitemsep,topsep=0pt,leftmargin=8pt]
        \item[\textcolor{cyan}{\dmark}] A Transformer-based \acs{DRL} method for solving \acf{EVRP}.
        \item[\textcolor{green}{\cmark}] Consider EV's energy and charging constraints. 
        \item[\textcolor{green}{\cmark}] Reduced energy consumption of \acs{EV} compared with exact algorithm, \acs{ACO}, \acs{ALNS}, and \acs{AM}.
        \item[\textcolor{red}{\xmark}] Lack model testing with real-world data.
        \item[\textcolor{red}{\xmark}] Need to consider data diversity, e.g., various \acs{EV} types.
    \end{itemize}
\end{minipage} \\ \hline
\end{tblr}
\end{table*}

\subsection{Layer 3: Smart Grid with \ac{EV}} 
The proliferation of \ac{EVs} leads to increased stochastic power demands from the grid, accelerating grid asset deterioration and complicating power system operations. The investigation of \ac{EV} charging load profiles is crucial for understanding future grid states to enable large-scale transportation electrification. However, the issues persist regarding the quantity and quality of data \cite{Ref21,Ref18}, and the challenges involved in predicting \ac{EV} charging loads \cite{Ref23,Ref31,Ref39,Ref28,Ref33} as well as the urge of understanding user experience \cite{Ref25}.

\subsubsection{Data Quantity and Quality}
The digitalization and widespread deployment of charging infrastructure offers a great opportunity to gather real \ac{EV} charging data, yet it is hampered by equipment failures, data collection errors, and intentional damage, resulting in missing values and outliers \cite{luo2018multivariate_R18r13}. Given insufficient data accumulation in newly built charging facilities, the \ac{ML} models can be biased with the flawed datasets \cite{gilanifar2021clustered_R18r14}, which are not necessary to be small-scale.
Such bias and inaccuracy pose challenges to the scheduling and optimization of the grid, whether centralized or distributed. As such, one important usage of \ac{GenAI} is to enhance the \ac{EV} and grid datasets to improve the system performance such as load forecasting and balancing. Several \ac{GenAI} algorithms, such as \ac{VAE} \cite{Ref21} and \ac{GAN} \cite{Ref17,Ref18,Ref33}, have been explored and we present the technical details of them below. 

\paragraph{\ac{VAE}-based Approach}
\ac{VAE}'s adoption is mainly for generating stochastic scenarios for \ac{EV} load profiles. In \cite{Ref21}, a \ac{VAE} model is designed for such usage to capture the time-varying and dynamic nature of \ac{EV} loads effectively. The paper considers five different \ac{EV} load profiles, for fully battery-based and hybrid-based \ac{EVs} with and without demand responses \cite{Ref21}. The profiles are measured at 10-minute intervals, resulting in 144 data points for each profile per day. The model uses the historical profiles as input, and accordingly generates new profiles that encapsulate the critical characteristics of the profiles. For future work, the proposed method holds the potential to be integrated with load forecasting models \cite{Ref31} as a data augmentation tool to address the data scarcity issue.

\paragraph{\ac{GAN}-based Approaches}
Similar to \ac{VAE}, \ac{GAN} has also been applied for data augmentation and furthermore, load forecasting. In \cite{Ref18}, \ac{GAN} is used to generate \ac{EV} charging load data first, with which, load forecasting is performed. The data augmentation model is called \acs{GRU-GAN}. The model uses the transactional data from various \ac{EV} charging stations within a 35 kV power distribution zone spanning from July 1 to August 31 in 2019 \cite{Ref18}. Due to different realistic restrictions, the time-series data is incomplete. As such, the data is pre-processed to simultaneously identify and handle missing values and outliers using \acs{GRU-GAN}, and this process is commonly referred to as data imputation. Finally, the generated high-quality data is used for training the Mogrifier \ac{LSTM} network for short-term \ac{EV} load forecasting. 

The comparison results show that \acs{GRU-GAN} outperforms conventional imputation techniques. Compared to mean imputation \cite{Book_mean_imputation}, \acs{GRU-GAN} can reduce the errors by 15\%. \acs{GRU-GAN}'s performance improvement is even more significant when comparing with piecewise linear \cite{web_PLI} and \ac{KNN} \cite{ZHANG20122541_KNN} imputation, achieving 43\% and 19\% lower errors, respectively. Such enhanced imputation accuracy is attributed to \acs{GRU-GAN}'s ability to capture intricate high-level data representations. The potential improvement of the solution is mainly in the forecasting stage. The forecasting of certain intervals is especially challenging, e.g., peak periods. The \ac{GenAI} models, e.g., transformer-based \cite{Ref23,Ref31}, may serve as viable substitutes for Mogrifier \ac{LSTM}.

\subsubsection{\ac{EV} Charging Load Prediction}
Understanding the pattern of \ac{EV} load is critical for the grid scheduling. Conventionally, the load prediction is formulated as a time series forecasting problem, which can be handled by different \ac{ML} algorithms such as the \ac{GenAI}-based ones. 

\paragraph{Traditional ML-based Approaches}
There exist several load forecasting approaches driven by traditional \ac{ML}, e.g., \ac{SVR} \cite{sun2016charging_R31r13}, \ac{ANN} \cite{mahmoud2015modelling_R31r14}, \acs{RNN} \cite{lai2018modeling_R31r15}, and \ac{LSTM} \cite{huang2020ensemble_R31r8}. These algorithms have demonstrated the potential of using ML for load forecasting. However, they either fell short in capturing the nonlinear characteristics inherent in \ac{EV} load series or faced challenges in modeling long-term dependencies that are common in real-world forecasting applications \cite{li2019enhancing_R31r17}. In the application layer, these algorithms focus on the load patterns in time scale where spatial load distribution is overlooked. A recent work \cite{LI2020_R39r24} developed a spatial–temporal forecasting method for load forecasting, but the method cannot model the correlation among spatial regions. 

To address the aforementioned challenges, the Transformer models \cite{Ref31,Ref23} and the \ac{GAN} models \cite{Ref39,Ref28,Ref33} have been introduced recently. The former utilizes attention mechanisms to capture the long-range dependencies and intricate patterns of the time-series data. This allows the model to focus on relevant parts of the input sequence and facilitate more accurate and robust load forecasting results. The latter helps to generate highly realistic \ac{EV} charging scenarios by learning the underlying data distribution over a latent space with conditions. A wide range of plausible scenarios generated by \ac{GAN} can provide a comprehensive view of potential future loads. We present the technical details as below.

\paragraph{Transformer-based Approach}
The transformer-based model has been widely used for time-series forecasting because it addresses the limitations of \ac{LSTM}s and effectively captures long-term dependencies \cite{Ref23}. The integration of a transformer-based architecture with the probabilistic forecasting technique in the temporal latent auto-encoder exhibited significantly enhanced efficacy in the context of time series forecasting endeavors \cite{nguyen2021AutoEncoder__R31r21}. Nevertheless, the vanilla transformer model exhibits significant time and memory overheads due to the quadratic computation complexity of self-attention, and imposes constraints on the maximum allowable input sequence length as a result of the accumulation of encoder and decoder layers \cite{zhou2021informer_R31r20}.  The Informer \cite{zhou2021informer_R31r20} tackles these challenges by substituting the conventional self-attention computation in the standard transformer with a ProbSparse self-attention framework \cite{zhou2021informer_R31r20}, while also introducing the self-attention distilling mechanism. However, it is limited to point forecasting which refers to the prediction of a single/specific value for a future variable or event \cite{web_PointForecast}.

The authors in \cite{Ref31} proposed a Probformer model to address the challenge of capturing long-term dependencies within charging load sequences and to facilitate the generation of probabilistic load forecasts. The model with multi-head ProbSparse self-attention \cite{zhou2021informer_R31r20} was an adaptation of the Transformer-based Informer framework \cite{zhou2021informer_R31r20}. Subsequently, the MetaProbformer \cite{Ref31} was developed by integrating the Probformer with a meta-learning algorithm, Reptile \cite{nichol2018firstorder_R31r37}, which tackled the issue of charging stations with scarce historical data. Four datasets comprising real-world \ac{EV} charging loads, sourced from various charging stations over distinct time periods, were utilized in the experiments. Each dataset details the start and end times, along with the total energy consumed (in kWh) for each charging event. While MetaProbformer is capable in load forecasting with limited historical data, it faces difficulties for new charging stations where data collection takes time. Hence, an exploration of methods for enhancing the model generalizability within a few-shot or zero-shot framework is necessary. 

\paragraph{GAN-based Approach}
Recently, the \ac{GAN}-based techniques for charging scenario generation have begun to emerge \cite{Ref39,Ref28,Ref33}. For example, the authors in \cite{Ref39} proposed \ac{WGAN-GP} to tackle spatial–temporal uncertainty in \ac{EV} charging load analysis. This approach explored load dynamics and generated scenarios without uniform probability assumptions across charging stations.
The undisclosed power grid structure limited data access to the \ac{EV} charging load. To model the impact of these stations on the distribution network, the data from 32 charging stations in the Zhejiang region were allocated to nodes in the IEEE 33-node distribution network system \cite{Ref39}. The forecasting target was the \ac{EV} charging load at each node in the network.
Nevertheless, the comparative analysis of the \ac{WGAN-GP} against contemporary \ac{GenAI} approaches in the context of \ac{EV} charging scenario generation remains lacking.

A \ac{CopulaGAN} model combining Copula transformation with \acs{GANs} was developed in \cite{Ref28}. Slightly different from the traditional load forecasting approaches focusing solely on \ac{EV} charging load curves, the \ac{CopulaGAN} captured uncertainties in \ac{EV} charging sessions, including energy delivered, arrival, and departure times. Subsequently, it was used for day-ahead optimal \ac{EV} scheduling. The charging session dataset used in this study was obtained from the Caltech parking lots \cite{CaltechParkData_R28r38}. Each site's database records the \ac{EV} connection time, charging completion time, energy consumption, and vehicle departure time from the parking lot. The power market dataset including the load forecast, generation forecast, day-ahead electricity market price, and balancing energy market, was collected from the German electricity market \cite{web_SMARD_German}. The simulation of \ac{EV} charging sessions and the prediction of day-ahead market prices were carried out for 48 hours. The outcomes of their investigation demonstrated a high degree of correspondence between the generated \ac{EV} charging session data and the actual dataset, with an approximate match rate exceeding 90\% \cite{Ref28}. This validation underscored the effectiveness of \ac{CopulaGAN} in accurately capturing the inherent structure and distributional attributes of the data. However, the proposed optimal scheduling of \ac{EVs} charging faces limitations due to data inadequacy and assumptions about \ac{SoC} and charging rates. The comparative analysis of \ac{CopulaGAN} and other \ac{GenAI} methods (e.g., GAN-DNN \cite{Ref27} and DiffCharge \cite{Ref26}) shall be investigated. 

\subsubsection{Consumer Data and Sentiment Analysis}
The transportation sector significantly contributes to \ac{GHG} emissions \cite{web_EPA_US} and impacts public health \cite{national2010hidden_R25r2}. Government policies increasingly favor \ac{EVs} to reduce \ac{GHG} emissions. However, the analysts underutilized consumer data, particularly unstructured \ac{EV} data, in decisions about charging infrastructure. Previous research employing sentiment analysis suggested prevalent negative user experiences in EV charging station reviews, yet lacked specific causal extraction \cite{asensio2020real_R25r11}. Thus, the multi-label topic classification is crucial for understanding user interaction behaviors in electric mobility. Hence, the authors in \cite{Ref25} utilized the Transformer-based models, \Ac{BERT} \cite{devlin2018bert_R25r26} and XLNet \cite{yang2019xlnet_R25r27}, for the multi-label topic classification in the domain of \ac{EV} charging reviews. The method proposed herein aimed to expedite the evaluation of research through automated means, utilizing extensive consumer data to assess performance and analyze regional policies. The study utilized data derived from 12,720 charging station locations across the United States, comprising 127,257 English-language consumer reviews written by 29,532 \ac{EV} drivers over a four-year period from 2011 to 2015 \cite{alvarez2019evaluating_R25r23}. An important future research is to enhance model interpretability through methods such as the use of rationales \cite{zaidan2008machine_R25r45}, influence functions \cite{serrano2019attention_R25r46}, and sequence tagging approaches \cite{nguyen2017aggregating_R25r47} to well understand consumers.

The \ac{GenAI} applications for IoEV in layer 3 are summarized in \cref{table:compare_layer3A}. We observed that the GAN models were often used for data augmentation \cite{Ref17,Ref33}, handling missing values and outliers of the input data \cite{Ref18}, and generating \ac{EV} charging sessions/scenarios \cite{Ref28,Ref39}. Besides \ac{GAN}, VAE can be utilized for generating stochastic scenarios for \ac{EV} load profiles. The Transformer-based models were usually utilized for the multi-label topic classification in the domain of \ac{EV} charging reviews \cite{Ref25}, and \ac{EV} load forecasting \cite{Ref31,Ref23}. 

\begin{table*} 
\caption{Summary of \ac{GenAI} for \ac{IoEV} in layer 3. \\ \textcolor{cyan}{\dmark}: \ac{GenAI} methods; \textcolor{green}{\cmark}: pros of the methods; \textcolor{red}{\xmark}: cons of the methods.}
\label{table:compare_layer3A} 
\begin{tblr}{
  width = \linewidth,
  colspec = {m{0.09\linewidth}m{0.07\linewidth}m{0.1\linewidth}m{0.635\linewidth}}, 
  cells = {c},      
  hlines,
  vline{2-4} = {-}{},
  vline{2} = {2}{-}{},
  hline{1,8} = {-}{0.08em},
}
\hline
\textbf{Applications} & 
\textbf{Reference} & 
\textbf{Techniques} & 
\textbf{Pros \& Cons} \\
%
\SetCell[r=2]{c} \acs{EV} Charging Load Profile Data Augmentation & 
\cite{Ref21} & 
\acs{VAE} & 
\begin{minipage}[c]{\linewidth}
    \begin{itemize}[noitemsep,topsep=0pt,leftmargin=8pt]
        \item[\textcolor{cyan}{\dmark}] A framework for generating and enhancing \acs{EV} load profiles.   
        \item[\textcolor{green}{\cmark}]  Ensure consistency in power consumption between generated and original profiles over time.  
        \item[\textcolor{green}{\cmark}] Capture temporal correlations, probability distributions, and volatility of original load profiles.   
        \item[\textcolor{red}{\xmark}] Challenge in balancing reconstruction loss and \acs{KL} divergence during training. 
    \end{itemize}
\end{minipage} \\
%
\acs{EV} Charging Load Profile Data Augmentation  & 
\cite{Ref18} & 
\acs{GRU-GAN} with Mogrifier \acs{LSTM} & 
\begin{minipage}[c]{\linewidth}
    \begin{itemize}[noitemsep,topsep=0pt,leftmargin=8pt]
        \item[\textcolor{cyan}{\dmark}]  A load data generation model for missing values and outliers of the input data. 
        \item[\textcolor{green}{\cmark}] Generate high-quality data closely resembling real data. 
        \item[\textcolor{green}{\cmark}] Outperform traditional mean \cite{Book_mean_imputation}, piecewise linear \cite{web_PLI}, and \acs{KNN} \cite{ZHANG20122541_KNN} imputation methods
        \item[\textcolor{red}{\xmark}] Sub-optimal forecasting performance during peak and plateau periods. 
    \end{itemize}
\end{minipage} \\
%
Analysis of EV Consumer &
\cite{Ref25} & 
\acs{BERT}, XLNet & 
\begin{minipage}[c]{\linewidth}
    \begin{itemize}[noitemsep,topsep=0pt,leftmargin=8pt]
        \item[\textcolor{cyan}{\dmark}] Transformer-based models for analyzing \acs{EV} charging reviews. 
        \item[\textcolor{green}{\cmark}] Outperform traditional \acs{LSTM} and \acs{CNN} models in terms of accuracy and F1 scores.
        \item[\textcolor{red}{\xmark}] Lack of interpretability. 
    \end{itemize}
\end{minipage} \\
%
\acs{EV} Load Forecasting & 
\cite{Ref31} & 
MetaProbformer & 
\begin{minipage}[c]{\linewidth}
    \begin{itemize}[noitemsep,topsep=0pt,leftmargin=8pt]
        \item[\textcolor{cyan}{\dmark}]  A Transformer-based model for \acs{EV} charging load forecasting.
        \item[\textcolor{green}{\cmark}] Perform well in point and probabilistic forecasting for the load at \acs{EV} charging station. 
        \item[\textcolor{green}{\cmark}] Competitive performance in probabilistic forecasting for both short-term and long-term tasks. 
        \item[\textcolor{green}{\cmark}]  Adaptable to seen and unseen scenarios. 
        \item[\textcolor{red}{\xmark}]    Need an extension to multivariate forecasting for future work. 
        \item[\textcolor{red}{\xmark}]  Need to improve model's generalization capabilities within a few-shot/zero-shot framework. 
    \end{itemize}
\end{minipage} \\
%
\SetCell[r=2]{c} Charging Scenarios Generation &
\cite{Ref39} &
\acs{WGAN-GP} &
\begin{minipage}[c]{\linewidth}
    \begin{itemize}[noitemsep,topsep=0pt,leftmargin=8pt]
        \item[\textcolor{cyan}{\dmark}]  A GAN-based model for generating \acs{EV} charging scenarios at charging stations.
        \item[\textcolor{green}{\cmark}] High degree of spatial correlation similarity in terms of \acs{SSIM} and \acs{FSIM} indexes.
        \item[\textcolor{red}{\xmark}]   Lack of comparative analysis for different \acs{GenAI} methods in \acs{EV} charging scenario generation.
    \end{itemize}
\end{minipage} \\
%
Charging Scenarios Generation &
\cite{Ref28} &
\acs{CopulaGAN} &
\begin{minipage}[c]{\linewidth}
    \begin{itemize}[noitemsep,topsep=0pt,leftmargin=8pt]
        \item[\textcolor{cyan}{\dmark}]     A GAN-based model for capturing and modelling the uncertainties in \acs{EV} charging sessions.
        \item[\textcolor{green}{\cmark}]   Generate \acs{EV} charging sessions data which closely align with the real dataset.
        \item[\textcolor{red}{\xmark}]  Real conditions not fully captured with simplified assumptions of \acs{SoC} and charging rate.
        \item[\textcolor{red}{\xmark}] Need a comparison study with other \acs{GenAI} approaches for scenarios generation.
    \end{itemize}
\end{minipage} \\\hline
\end{tblr}
\end{table*}

\subsection{Layer 4: Security}
Security is a critical aspect that we cannot bypass for many cyber-physical systems. In the following parts, we introduce the background of security research for electric mobility and then present related \ac{GenAI} research works.

\subsubsection{Background}
Security's impact on electric mobility encompasses several aspects. The first one is due to the ever-increasing adoption of the \ac{ML} models. For example, there has been considerable utilization of \ac{DRL} algorithms within the context of \ac{EV} charging schedules, aiming at acquiring optimal charging strategies for users. Such algorithms include \ac{SAC} \cite{Ref7}, \ac{DDPG} \cite{Ref8}, \ac{SDRL} \cite{Ref9}, \ac{MADRL} \cite{Ref14}, \ac{PPO} \cite{Ref13}, etc. Same as many \ac{ML} algorithms, \ac{DRL} suffers from potential adversarial attacks \cite{wei2023self_R36r9}. The adversarial examples involve maliciously altered inputs to deceive the \ac{ML} models, causing erroneous outputs \cite{chakraborty2021survey_R36r26}. Some methods are shown to be effective in generating adversarial data, e.g., \ac{FGSM} \cite{huang2017adversarial_R36r8}, \ac{BIM} \cite{kurakin2018_R36r12}, and DeepFool \cite{moosavi2016deepfool_R36r28}. The ramifications of such attacks could be extensive, ranging from increasing charging expenses and fluctuations in grid loads to jeopardizing the stability of power grids \cite{wang2020adversarial_R36r13}. 

Besides adversarial attacks, security is also a concern in power grids with \ac{EV} penetration. Specifically, the integration of power systems with electrified transportation networks poses new challenges concerning the reliability and resilience of charging infrastructure \cite{gan2021tri_R34r3}. According to \cite{lu2022multi_R34r5}, there is an increase in the occurrence of power system attacks targeting customer satisfaction levels in charging services. Deep learning algorithms such as \ac{DBN} \cite{li2018false_R34r20} and \ac{FNN} \cite{xue2019detection_R34r16} have been used for detecting cyber-physical attacks in power systems because of their advanced feature extraction capabilities. The algorithms achieve high detection rates, yet they overlook the extraction of the important spatial relationships inherent in the data, as they disregard the topological grid attributes \cite{boyaci2021graph_R34r24}. 

The connection between \ac{EV}s and the grid is largely controlled by the \ac{EV}'s supply equipment, that is responsible for managing and maintaining the charging operations. It also facilitates communication among cloud services, payment providers, \ac{EVs}, battery management systems, and other relevant entities to enable efficient and intelligent charging \cite{en15113931_R44r4}. However, the connection is often associated with potential vulnerabilities such as \ac{DoS} attacks, \ac{FDIAs}, and spoofing \cite{johnson2022cybersecurity_R44r5}. A popular connection standard is \ac{CAN}, which is a bus protocol for communication within vehicles. \ac{CAN} has many advantages such as reduced wiring expenses, minimal weight, and simplified design \cite{lee2017otids_R47r21}. However, \ac{CAN} has security vulnerabilities also, including inadequate authentication mechanisms, susceptibility to multiple attack vectors, and the absence of encryption technologies \cite{Verma2020ROADTR_R47r22}. The security vulnerabilities have been studied with several methods such as \ac{LOF}, \ac{OCSVM}, and \ac{PCA} \cite{marino2019cyber_R44r9}. Nevertheless, these methods often are sensitive to noise, incapable to handle high-dimensional data, and fail to handle the intricate dynamics of systems.

\subsubsection{Adversarial Attacks} 
To investigate adversarial attacks against \ac{DRL} in the \ac{EV} charging process, a GAN-based approach namely \acs{RL-AdvGAN} was introduced in \cite{Ref36}. \Acs{RL-AdvGAN} could support the adversary to leverage the stolen data to engage in behavior cloning, thereby constructing an adversarial policy network to mimic the user's policy. In this research, the \ac{EV} charging environment was set up using real-world data from California \ac{ISO} \cite{web_CaliforniaISO}. The dataset spanned a duration of approximately 2 years with data recorded hourly, consisting of the electricity prices and the grid load information. The paper assumed that the commuting behavior of \ac{EV} users follows the normal distributions with parameters of arrival time, departure time, and battery \ac{SoC}. Four types of \ac{DRL} algorithms, i.e., \ac{DQN} \cite{mnih2015human_R36r33}, \ac{DDPG} \cite{lillicrap2019_R36r34}, \ac{PPO} \cite{schulman2017proximal_R36r35}, and \ac{SAC} \cite{haarnoja2018soft_R36r36} were tested under the adversarial attacks generated by \ac{FGSM} \cite{goodfellow2014explaining_R36r11} and \acs{RL-AdvGAN} \cite{Ref36}. According to the results, \ac{FGSM} attack is effective, and the algorithms, i.e., \ac{DQN}, \ac{DDPG}, \ac{PPO}, and \ac{SAC}, only managed to maintain a normal \ac{SoC} range of the battery in 6\%, 0\%, 28\%, and 8\%, of the time, respectively \cite{Ref36}. The proposed \acs{RL-AdvGAN} attack \cite{Ref36} was even more effective, where all the tested \ac{DRL} algorithms failed to maintain the normal range of \ac{SoC} all the time during the attack. As such the proposed \acs{RL-AdvGAN} posed a significant risk to the security and stability of the battery system. Future work includes further exploring the threat and damage of \ac{GenAI} methods for adversarial attacks with different network architectures and developing effective attack detection methods.

In \cite{Ref45}, the authors considered both the generation and detection of adversarial attacks against \ac{V2M} systems and proposed GAN-based models. The research modeled a system where the adversaries sought to manipulate the \ac{ML} classifier at the network edge, causing it to incorrectly classify the energy requests received from microgrid users, e.g. \ac{EVs}' charging/discharging requests. The \ac{FGSM} and \ac{CGAN} models were introduced to serve as attackers in generating adversarial instances. On the adversarial detection side, a GAN-based adversarial training framework was introduced to create adversarial training instances. The instances are used to train \ac{SVM} classifiers to detect these adversarial attacks. The simulation was set up using the iHomeLab RAPT dataset \cite{web_data_2019} consisting of electrical power consumption and generation of 5 Swiss households, and the supplementary dataset with power generation from the batteries and wind farm from their previous work \cite{omara2021impact_R45r12}. The results indicate that their proposed method outperforms the \ac{DBSCAN} algorithm, leading to an improvement in the adversarial detection rate ranging from 13.2\% to 25.6\% \cite{Ref45}. For future work, the investigation of the impact of limited resources on the adversarial detection rate is necessary since the classifier is designed to be deployed at the network edge.

\subsubsection{Detection of Attacks}
The authors in \cite{Ref34} extended the study of potential attacks \cite{Ref36} and developed a \ac{GAE} model for detecting the \ac{FDIAs} in power systems. The model leverages the correlations between power system data (e.g., active and reactive power measurements) and transportation data (e.g., hourly traffic volume) to enhance charging satisfaction. In the event of \ac{FDIAs}, the malicious entities can manipulate power measurements to simulate additive, deductive, and camouflage attacks. The level of satisfaction for customers can diminish due to insufficient power supply at the charging stations, e.g., blocked charging requests or extended charging times. The proposed method was tested via simulation of the Texas power grid consisting of 2,000 buses and 360 active charging stations. The load data was sourced from \ac{ERCOT} which manages the distribution of electric power to more than 27 million customers in Texas \cite{web_ercot_data}. Compared to the benchmark graph \ac{CNN} model \cite{boyaci2021graph_R34r24}, the proposed \ac{GAE} model improved the detection accuracy by about 15\% in various \ac{FDIAs} scenarios on vulnerable nodes when 30\% of data was under attack. The results show that the proposed \ac{GAE} model outperforms the state-of-the-art detector in various attacks. For example, compared to the benchmark graph \ac{CNN} model \cite{boyaci2021graph_R34r24}, the proposed \ac{GAE} model can improve the detection accuracy by around 14.4\% to 14.8\% in the various \ac{FDIAs} scenarios e.g, additive attacks, deductive attacks, and combined attacks \cite{Ref34}. Future research may focus on real-time model updates and decision-making in dynamic operational contexts of power and transportation networks.

Based on the above introduced detection model, the researchers in \cite{Ref47} developed an interpretable anomaly detection system, referred to as RX-ADS. The system was designed to identify intrusions within the \ac{CAN} protocol communications of the \ac{EV}s with active charging connections. Besides, the ResNet Autoencoder was employed to learn normal behavior from data and detect anomalies based on reconstruction errors. The performance of the system was investigated with two publicly available datasets of \ac{EV} \ac{CAN} protocol, including \ac{OTIDS} \cite{web_Ref47r21_OTIDS_dataset} and Car-Hacking \cite{web_Ref47r10_CarHacking_dataset}. The results showed that the system outperformed a GAN-based intrusion detection system \cite{seo2018gids_R47r10} by 4\% under \ac{DoS} attacks. Nevertheless, the proposed method requires a substantial dataset that accurately reflects the system's typical normal behavior. If new behaviors arise that deviate from the established norms, the model shall be updated. Otherwise, such deviations might be incorrectly classified as anomalies, leading to an increase in false positives.

The same research group of \cite{Ref47} developed a ResNet AE-based approach for unsupervised physical anomaly detection with high resilience in \cite{Ref48}, specifically in \ac{EV} charging stations without labeled data. The experiments were conducted using data under normal and physical attack scenarios from the Idaho National Laboratory's \ac{EV} charging station system testbed. The proposed ResNet Autoencoder based approach was compared with two benchmark algorithms: \ac{LOF} and \ac{OCSVM}. The results in terms of F1 score show that the proposed method is able to improve the detection performances of \ac{LOF} algorithm by around 20.1\% and \ac{OCSVM} algorithm by about 3.7\% \cite{Ref48}. For further work, enhancing the proposed anomaly detection framework can be achieved by integrating cyber security related scenarios pertinent to \ac{EV} charging systems.

In addition to detecting \ac{FDIAs} \cite{Ref34,Ref46}, the authors in \cite{Ref44} proposed a ResNet AE-based anomaly detection framework which consists of \ac{Cy-ADS} and \ac{Phy-ADS} for the cyber and physical data streams, respectively. The former has the capability to monitor and analyze packet data in real-time for the detection of cyber anomalous behaviors within \ac{EV} charging stations. Meanwhile, the latter is designed to capture and analyze real-time physical sensor data (e.g., voltage, current, power, and thermal measurements) to identify physical anomalous behavior. Their results show that the proposed \ac{Cy-ADS} for detection of cyber attack and \ac{Phy-ADS} for physical anomaly detection outperforms the \ac{LSTM} \ac{AE} method by about 18\% and 15\% respectively, \ac{VAE} approach by around 3\% and 5\% respectively \cite{Ref44}. However, the proposed method is subject to the quality of the training data and model retraining is needed if system behaviors change over time. 

The \ac{GenAI} research for \ac{EV} and \ac{IoEV} security in layer 4 is summarized in \cref{table:compare_layer5}. As seen from the table, \ac{AE}-based and \ac{GAN}-based models are commonly found in recent research for security. Among them, GAN-based models are mainly used for generating and identifying adversarial attacks, e.g., \ac{RL-AdvGAN} \cite{Ref36} and \ac{CGAN} \cite{Ref45}. Whereas the AE-based models \cite{Ref34,Ref44,Ref47,Ref48} are good at detecting \ac{FDIAs}, \ac{DoS}, fuzzy, and physical attacks, and serve as viable alternatives to traditional methods such as \ac{LOF} and \ac{OCSVM}.

\begin{table*} 
\caption{Summary of \ac{GenAI} for \ac{IoEV} in layer 4. \\ \textcolor{cyan}{\dmark}: \ac{GenAI} methods; \textcolor{green}{\cmark}: pros of the methods; \textcolor{red}{\xmark}: cons of the methods.}
\label{table:compare_layer5} 
\begin{tblr}{
  width = \linewidth,
  colspec = {m{0.09\linewidth}m{0.07\linewidth}m{0.1\linewidth}m{0.635\linewidth}}, 
  cells = {c},      
  hlines,
  vline{2-4} = {-}{},
  vline{2} = {2}{-}{},
  hline{1,8} = {-}{0.08em},
}
\hline
\textbf{Applications} & 
\textbf{Reference} & 
\textbf{Techniques} & 
\textbf{Pros \& Cons} \\
%
\SetCell[r=2]{c} Adversarial Attacks & 
\cite{Ref36} & 
\acs{RL-AdvGAN} & 
\begin{minipage}[c]{\linewidth}
    \begin{itemize}[noitemsep,topsep=0pt,leftmargin=8pt]
        \item[\textcolor{cyan}{\dmark}] A GAN-based model for generating adversarial attacks against \acs{DRL} algorithms. 
        \item[\textcolor{green}{\cmark}] Effectively reduce the performance of common \acs{DRL} algorithms for optimal \acs{EV} charging. 
        \item[\textcolor{green}{\cmark}] Higher security threat than \acs{FGSM} \cite{huang2017adversarial_R36r8}.
        \item[\textcolor{red}{\xmark}]  Need to improve training stability with model refinement.
        \item[\textcolor{red}{\xmark}] Need to propose methods to prevent adversarial attacks.
    \end{itemize}
\end{minipage} \\
%
Adversarial Attacks & 
\cite{Ref45} & 
\acs{GAN} and \acs{CGAN} & 
\begin{minipage}[c]{\linewidth}
    \begin{itemize}[noitemsep,topsep=0pt,leftmargin=8pt]
        \item[\textcolor{cyan}{\dmark}]   A GAN-based detection framework for identifying the adversarial attacks. 
        \item[\textcolor{green}{\cmark}]  Higher adversarial detection rate compared with traditional \acs{DBSCAN} algorithm \cite{Ref45}. 
        \item[\textcolor{red}{\xmark}]  Incompatible with network edge services due to resource constraints. 
    \end{itemize}
\end{minipage} \\
%
False Data Injection Attacks & 
\cite{Ref34} & 
Graph Autoencoder & 
\begin{minipage}[c]{\linewidth}
    \begin{itemize}[noitemsep,topsep=0pt,leftmargin=8pt]
        \item[\textcolor{cyan}{\dmark}]   An AE-based detection scheme for identifying \acs{FDIAs}.  
        \item[\textcolor{green}{\cmark}]  Performance improvement of 15-25\% compared with \acs{SVM}, \acs{FNN}, \acs{CNN}, and \acs{LSTM} \cite{Ref34}. 
        \item[\textcolor{red}{\xmark}]  Need to update and deploy the offline-trained model to real-time applications for future work. 
    \end{itemize}
\end{minipage}  \\
%
Cyber and Physical Attacks & 
\cite{Ref44} & 
ResNet Autoencoder (AE) & 
\begin{minipage}[c]{\linewidth}
    \begin{itemize}[noitemsep,topsep=0pt,leftmargin=8pt]
        \item[\textcolor{cyan}{\dmark}]   An AE-based anomaly detection framework for identifying both cyber and physical attacks. 
        \item[\textcolor{green}{\cmark}] Capable of detecting both simple and complex cyber-physical attack scenarios.   
        \item[\textcolor{green}{\cmark}]  Low training and inference time, making it suitable for real-time applications. 
        \item[\textcolor{red}{\xmark}]  Sensitive to the setting of threshold values.  
    \end{itemize}
\end{minipage} \\
%
\Acs{DoS} and Fuzzy Attacks  & 
\cite{Ref47} & 
ResNet Autoencoder (AE) & 
\begin{minipage}[c]{\linewidth}
    \begin{itemize}[noitemsep,topsep=0pt,leftmargin=8pt]
        \item[\textcolor{cyan}{\dmark}]  An AE-based method for detecting intrusions in \acs{CAN} communication protocol for \acs{EV} charging.
        \item[\textcolor{green}{\cmark}]  Competitive results compared with the GAN-based approach \cite{seo2018gids_R47r10}.
        \item[\textcolor{red}{\xmark}] Require a large amount of data that reflects the normal behavior of the system. 
        \item[\textcolor{red}{\xmark}] Need model update and retraining for new behaviors in the future. 
    \end{itemize}
\end{minipage} \\
%
Physical Attacks & 
\cite{Ref48} & 
ResNet Autoencoder (AE) & 
\begin{minipage}[c]{\linewidth}
    \begin{itemize}[noitemsep,topsep=0pt,leftmargin=8pt]
        \item[\textcolor{cyan}{\dmark}]    An AE-based approach for physical anomaly detection. 
        \item[\textcolor{green}{\cmark}] Outperform two benchmark algorithms including \acs{LOF} and \acs{OCSVM} \cite{Ref48}.  
        \item[\textcolor{red}{\xmark}]  Not capable of detecting cyber attacks.
    \end{itemize}
\end{minipage} \\ \hline
\end{tblr}
\end{table*}

\subsection{Studies Across Multiple Layers}  

Besides studies on individual layer, some studies were carried out on multiple \ac{IoEV} layers, e.g., \cite{Ref26,Ref29}. Understanding user behavior \cite{li2018gis_R29_user_behavior_analysis}, coupling characteristics \cite{liu2022fast_R29_coupling} among user behavior, road networks \cite{arias2017_R29_road_network}, and \ac{EVs} are crucial for accurate demand prediction, but it remains challenging due to various factors such as time and \ac{SoC}. Nevertheless, research on precise mathematical models for charging and discharging strategies is lacking due to the complexity of influencing factors.    

\paragraph{Generation of \ac{EV} Charging Scenarios}
In \cite{Ref26}, a diffusion model namely DiffCharge was developed to generate \ac{EV} charging scenarios. The generated scenarios could be divided into battery-level (e.g., charging current in Ampere) and station-level (e.g., charging load in kW). The traditional \ac{ML} e.g., \ac{GMM} could be utilized to estimate the daily \ac{EV} charging load profiles, but it faced a challenge in accurately capturing temporal dynamics across various time-series \ac{EV} charging data. On the other hand, DiffCharge as one of the diffusion-based approaches is capable of deriving the challenging uncertainties associated with charging and producing a range of charging load profiles characterized by realistic and unique temporal features. On top of \ac{DDPM} \cite{ho2020denoising}, the DiffCharge framework consists of \ac{LSTM}, broadcast, multi-head self-attention, and 1D-CNN. DiffCharge was trained by using the ACN-Data \cite{CaltechParkData_R28r38} dataset which comprises real-world charging data of individual \ac{EVs} in California. The data attributes including connection time, done charging time, kWh delivered, and charging current in Ampere were considered for training the model to generate \ac{EV} charging curves. The daily charging load profile could be aggregated and extracted from the ACN-Data dataset. Then, it was integrated with arrival/departure time and actual scheduled energy for training the model to generate the \ac{EV} load profile at the station. The proposed method was compared to the baseline models: \ac{GMM} \cite{powell2022large}, VAEGAN \cite{de2021unsupervised}, and TimeGAN \cite{yoon2019time}. Their results showed that the proposed DiffCharge could generate realistic charging curves and it outperformed the baseline models in terms of marginal score, discriminative score, and tail score. For example, the marginal score of DiffCharge was improved by 91\% from \ac{GMM}, by 35\% from VAEGAN, and by 32\% from TimeGAN \cite{Ref26}. However, the control capability of the DiffCharge is restricted, thereby limiting the model's ability to generate tailored charging scenarios for varying conditions such as initial \ac{SoC}, types of batteries, and station congestion.

\paragraph{Generation of Regional Electric Vehicle (EV) Charging Demand}
Similar to \cite{Ref26}, \cite{Ref29} explored the \ac{EV} charging demand from perspectives encompassing both battery and station levels. However, a notable distinction of \cite{Ref29} lies in the emphasis placed on the spatial–temporal distribution of \ac{EV} charging demand in the region. The authors in \cite{Ref29} addressed the challenges of predicting spatial–temporal \ac{EV} charging demand at both battery and station levels by proposing a deep learning framework that consists of methodologies such as \ac{GAIL}, \ac{PPO}, and XGBoost. The paper classified strategies concerning user charging and discharging into three categories: driving policies, travel target mileage policies, and charging duration selection policies. Following this categorization, the research leveraged \ac{GAIL} to obtain insights into these delineated policies i.e., \ac{GAIL} was used as a strategy learning model. Then, employing the \ac{PPO}, the strategy learning model underwent optimization utilizing an \ac{SoC} forecasted through the XGBoost algorithm. The data utilized in this study were acquired from the Shanghai New Energy Electric Vehicle Monitoring Center \cite{web_Shanghai_data}, pertaining to a cohort of 1,000 \ac{EVs} subjected to testing over the course of one month. The data attributes included speed, acceleration, \ac{SoC}, temperature, longitudes, and latitudes. The data points were sampled every 10 seconds. The output variables were the 24-hour \ac{SoC} predictions for individual vehicles and the forecast of regional spatial–temporal charging demand. Four categories of vehicle's \ac{SoC}, encompassing logistics vehicles, taxis, buses, and private cars, underwent predictive analysis. The assessment criteria exhibited a variability spanning approximately 1.66\% to 3.15\% for \ac{MAE} and 2.32\% to 4.44\% for \ac{RMSE} \cite{Ref29}. Future research will refine the findings of this study. Considering additional factors such as road conditions and user demographics will enhance the accuracy of \ac{EV} \ac{SoC} predictions. 

\Cref{table:compare_multilayer} summarizes \ac{GenAI} implementations for \ac{IoEV} applications across multiple layers. We discovered that a few papers considered both layer 1 and layer 3 applications. For example, DiffCharge \cite{Ref26} is able to generate \ac{EV} charging scenarios for both battery level and EV charging station level. In contrast, in  \cite{Ref29}, \ac{SoC} prediction and regional charging load forecasting were completed by a framework with \ac{GAIL}.

\begin{table*}
\caption{Summary of \ac{GenAI} for \ac{IoEV} across multiple layers. \\ \textcolor{cyan}{\dmark}: \ac{GenAI} methods; \textcolor{green}{\cmark}: pros of the methods; \textcolor{red}{\xmark}: cons of the methods.}
\label{table:compare_multilayer} 
\begin{tblr}{
  width = \linewidth,
  colspec = {m{0.09\linewidth}m{0.07\linewidth}m{0.1\linewidth}m{0.635\linewidth}}, 
  cells = {c},      
  hlines,
  vline{2-4} = {-}{},
  vline{2} = {2}{-}{},
  hline{1,8} = {-}{0.08em},
}
\hline
\textbf{Applications} & 
\textbf{Reference} & 
\textbf{Techniques} & 
\textbf{Pros \& Cons} \\
%
Scenarios Generation    &
\cite{Ref26} & 
DiffCharge & 
\begin{minipage}[c]{\linewidth}
    \begin{itemize}[noitemsep,topsep=0pt,leftmargin=8pt]
        \item[\textcolor{cyan}{\dmark}]   A diffusion-based model for generating \acs{EV} charging scenarios for both battery-level (Layer 1) and station-level (Layer 3).    
        \item[\textcolor{green}{\cmark}]   Outperform \acs{GMM} \cite{powell2022large}, VAEGAN \cite{de2021unsupervised}, and TimeGAN \cite{yoon2019time} in charging scenarios generation. 
        \item[\textcolor{red}{\xmark}]    Limited control over diverse conditions, e.g., initial \acs{SoC}, battery types, and station congestion.  
    \end{itemize}
\end{minipage} \\
%
\acs{SoC} and Load Forecasting &
\cite{Ref29} & 
\acs{GAIL}, \acs{PPO}, and XGBoost \cite{Ref29} & 
\begin{minipage}[c]{\linewidth}
    \begin{itemize}[noitemsep,topsep=0pt,leftmargin=8pt]
        \item[\textcolor{cyan}{\dmark}]  A GAN-based model, \acs{GAIL}, as a strategy learning model assisting \acs{DRL} algorithm. 
        \item[\textcolor{green}{\cmark}]   Good \acs{SoC} prediction with low \acs{MAE} and \acs{RMSE} values. 
        \item[\textcolor{red}{\xmark}]   Overlook predictions at the individual charging station level.    
    \end{itemize}
\end{minipage} 
\\\hline
\end{tblr}
\end{table*}

\subsection{Discussion}
We have presented technical details of GenAI's usage in IoEV in different layers in various aspects and we present our discussions about GenAI's performance below.

\subsubsection{Key Features for GenAI's Competitiveness} 
We may notice that GenAI is not the first ML technology but manages to outperform traditional ML algorithms for many discussed applications and tasks. We summarize and highlight the key features that drive GenAI's competitiveness in IoEV.

\paragraph{Data Generation and Robustness} 
Data scarcity has been among the toughest challenges for traditional ML, e.g., in layers 2 and 3 for predicting supply and demand. GenAI algorithms such as GAN and VAE help generate realistic synthetic data by learning the underlying distribution of the limited raw data. Such GenAI algorithms demonstrate a high level of noise control during data generation and this enhances the stability of anomaly detection, predictive modeling, etc.

\paragraph{Advanced Pattern Recognition}
Compared to traditional ML, GenAI exhibits improved pattern recognition performance, e.g., for modeling and prediction tasks. This is critical for IoEV-related applications that involve complex (e.g., high-dimensional and nonlinear) patterns and dependencies. Two representative techniques are GDM and transformer. The former captures inherent distribution characteristics and models complex relationships to understand system dynamics. The latter extracts multi-scale features and exploits long-term dependencies well with its self-attention mechanism. 

With these strengths, GenAI becomes a versatile and competitive solution, and we foresee GenAI advancements and its increased usage in IoEV applications in the future.

\subsubsection{Selection of GenAI Algorithms} 
GenAI has been used for various IoEV applications and the optimal GenAI performance in part depends on the choice of GenAI algorithms. Each GenAI algorithm owns unique characteristics that can influence the algorithm's performance in solving different problems. For example, GAN has been shown to be a popular technology for data augmentation by generating high-fidelity synthetic data (e.g., charging behavior). When robustness is a key concern, GAN becomes less competitive due to its ineffectiveness in capturing data diversity well and mode collapse issue. VAE is capable of probabilistic data generation and particularly useful in generating stochastic scenarios. However, its reliance on the Gaussian latent space sacrifices data details sometimes. The diffusion model outperforms GAN for producing high-resolution outputs by iteratively refining noisy data, but its demand for computing resources is significant. Overall, we urge a comprehensive evaluation of different GenAI algorithms and the selection of suitable algorithms to meet specific requirements and constraints in different tasks.

\section{Technical Reviews: Dataset}
\label{sect4:GAI_IoEV_Dataset}
Data is highly important for \ac{GenAI} for model training, system customization, performance improvement, and so on. In this section, we provide a summary of the available public dataset in the domain of \ac{GenAI}-based electric mobility applications. We summarize the datasets in \cref{table:compare_dataset} and describe them in detail as below.

\begin{table*} 
\centering
\caption{Summary of dataset used in \ac{GenAI} for \ac{IoEV} at different layers.}
\label{table:compare_dataset}
\begin{tblr}{ width = \linewidth,
  colspec = {Q[1.5,c,m] Q[1.5,c,m] Q[5,l,m] Q[3.2,l,m] Q[1.9,c,m]},
  row{1} = {c}, 
  hlines,
  vline{2-5} = {-}{},
  vline{2} = {2}{-}{},
}
\hline
\textbf{Applications} & \textbf{Dataset} & \textbf{Properties} & \textbf{Dataset Size} & \textbf{Algorithms} \\
\SetCell[c=5]{c} Layer 1\\
\SetCell[r=2]{c} Anomaly Detection &
\ac{NSMC-EV} \cite{li2020battery}   &
13-dimensional time series, e.g., vehicle speed, charging state, insulation resistance, and \ac{SoC}. &      
\Ac{NSMC-EV} platform for over three million \ac{EVs}  &
\acs{GRU-VAE} \cite{Ref20} \\ 
%
Anomaly Detection &
Faults and Failure \cite{zhao2022data_R41r41} &
Multiple scenarios, e.g., short circuit and thermal runaway.
Time series of voltage, current, etc.
&                       
316 \ac{NCM} battery cells \cite{Ref41} &
\Ac{BERTtery} \cite{Ref41} \\
%
%
\SetCell[r=2]{c} \Ac{SoC} Estimation &
\ac{EV} \cite{en14123692} &
Time series, e.g., vehicle speed, voltage, cell temperature, motor controller voltage, and SoC. &
%
Driving data from five identical vehicles over a year \cite{en14123692}  &
\ac{TS-WGAN} \cite{Ref30} \\
%
\Ac{SoC} Estimation &
%
Li-ion Battery \cite{8790543} &
Time series of recorded variables, e.g., cell voltage, current, battery temperature, and ampere-hours. &  
%
Various drive cycles, including US06, UDDS, and LA92. &
%
\ac{TS-WGAN} \cite{Ref30}  \\
%
\Ac{SoH} Estimation &
LFP Battery \cite{RN713}   &
Rated capacity, number of cells, charging current, discharging current, cut-off voltage, etc.  &   
4 cells; cycles: 1062, 1266, 1114, and 1047. \cite{Ref37} &
\ac{DDPM} \cite{Ref37}  \\
\end{tblr}

\begin{tblr}{ width = \linewidth,
  colspec = {Q[1.5,c,m] Q[1.5,c,m] Q[5,l,m] Q[3.2,l,m] Q[1.9,c,m]},
  row{1} = {c}, 
  hlines,
  vline{2-5} = {-}{},
  vline{2} = {2}{-}{},
}
\SetCell[c=5]{c} Layer 2\\
\ac{EV} Charging Behaviors &
%
EA Technology \cite{web_dataset_Ref27} &
Residential charging events, e.g., date time, arrival hour, plug-out hour, and required energy  &  
Charging behaviors of over 200 participants in 2014 and 2015. 
&
%
\ac{GAN} \cite{Ref27} \\
%
%
%
Smart Home &
iHomeLab PART \cite{huber2020residential_R35r27} &
Energy consumption of households and specific appliances as well as \ac{PV} generation.  &
%
Five houses in the Lucerne region, Switzerland. &
%
\acs{VAE-GAN} \cite{Ref35}, \acs{GANs} \cite{Ref45} \\
%
%
\SetCell[r=2]{c} Residential EV Load &
\ac{CLNR} TC1a \cite{web_data_R43r1} &
UK electricity customers’ electricity consumption measured by British Gas's smart meter;   &     
Up to 8,000 customers for the year 2011.  &
\ac{GAN} \cite{Ref43}  \\
& \ac{CLNR} TC5 \cite{web_data_R43r2} & TC5 \cite{web_data_R43r2}: including customers’ energy use and solar PV performance. & Part of the project involving over 12,000 consumers. & \ac{GAN} \cite{Ref43} \\
\end{tblr}

\begin{tblr}{ width = \linewidth,
  colspec = {Q[1.5,c,m] Q[1.5,c,m] Q[5,l,m] Q[3.2,l,m] Q[1.9,c,m]},
  row{1} = {c}, 
  hlines,
  vline{2-5} = {-}{},
  vline{2} = {2}{-}{},
}
\SetCell[c=5]{c} Layer 3\\
\ac{EV} Load Forecasting &
City of Boulder \cite{web_colorado_OpenData_gov} &
Charging load records, e.g., address, arrival and departure time, type of the plug,  data, and energy. &                       
4 years; 25 public charging stations in Boulder. &
%
Transformer \cite{Ref23} \\
%
%
%
\SetCell[r=2]{c} Scenarios Generation &
%
\acs{ACN} Data \cite{CaltechParkData_R28r38} &
Time of \ac{EV} connection, charging time, amount of energy received by \ac{EV}, and time of leaving. &  
%
Over 30,000 charging sessions; growing daily \cite{CaltechParkData_R28r38}. &
\Ac{CopulaGAN} \cite{Ref28}; DiffCharge \cite{Ref26} \\
%
%
%
%
Scenarios Generation &
Belgian Elia Group \cite{web_elia_data} &
%
\ac{PV} and load power in the microgrid.  &                       
2 months of 15-minute interval data. &
%
\Ac{GAN} \cite{Ref33} \\
\end{tblr}
\begin{tblr}{ width = \linewidth,
  colspec = {Q[1.5,c,m] Q[1.5,c,m] Q[5,l,m] Q[3.2,l,m] Q[1.9,c,m]},
  row{1} = {c}, 
  hlines,
  vline{2-5} = {-}{},
  vline{2} = {2}{-}{},
}
\SetCell[c=5]{c} Layer 4\\
Adversarial Attacks &
California \acs{OASIS} \cite{web_CaliforniaISO} &
Electricity prices and the grid load information from OASIS site. &   
Continuously updating \cite{web_CaliforniaISO}  & 
%
\acs{RL-AdvGAN} \cite{Ref36} \\
%
%
%
\ac{FDIAs} &
\ac{ERCOT}'s Data \cite{web_ercot_data} &
Load profiles from the grid. &                       
Continuously updating \cite{web_ercot_data}. &
%
\Ac{GAE} \cite{Ref34} \\
%
%
\SetCell[r=2]{c} \acs{DoS} and Fuzzy Attacks &
\ac{OTIDS} dataset \cite{web_Ref47r21_OTIDS_dataset} &
States of \ac{DoS} attacks, fuzzy attacks, impersonation attacks, and attack-free conditions. 
&      
%
Attacks: 657K \ac{DoS}, 592K fuzzy, and 995K impersonation. & 
%
RX-ADS \cite{Ref47} \\
%
 &
Car-Hacking \cite{web_Ref47r10_CarHacking_dataset} &
Attack types, e.g., \Ac{DoS}, fuzzy, drive gear spoofing, and \acf{RPM} gauge spoofing. 
&  
%
Attacks: 3.6M \ac{DoS}, 3.8M fuzzy, 4.4M spoofing
drive gear, etc. & 
%
RX-ADS \cite{Ref47} \\
\end{tblr}
\begin{tblr}{ width = \linewidth,
  colspec = {Q[1.5,c,m] Q[1.5,c,m] Q[5,l,m] Q[3.2,l,m] Q[1.9,c,m]},
  row{1} = {c}, 
  hlines,
  vline{2-5} = {-}{},
  vline{2} = {2}{-}{},
}
\SetCell[c=5]{c} Multiple Layers\\
\Ac{SoC} and Load Forecasting &
Shanghai New Energy \cite{web_Shanghai_data} &
Data sampled for speed, acceleration, \ac{SoC}, temperature, longitudes, and latitudes. &  
%
Continuously updating \cite{web_Shanghai_data}  &
%
\Ac{GAIL}, \ac{PPO}, XGBoost \cite{Ref29} \\
\hline
\end{tblr}
\end{table*}

\textit{Layer 1:}
In the battery layer, the dataset from \ac{NSMC-EV} \cite{li2020battery} and the unlabeled dataset on multiple faults and failure scenarios \cite{zhao2022data_R41r41} are used by \acs{GRU-VAE} \cite{Ref20} and \ac{BERTtery} \cite{Ref41}, respectively, for the anomaly detection. \Ac{NSMC-EV} serves as China’s national big data platform for EVs, offering extensive real-time online data on EVs utilized in public transportation. The dataset used for \acs{EV}'s \ac{SoC} estimation can be found in \cite{en14123692}, \cite{8790543}, which provides the battery's voltage, current, temperature, \ac{SoC}, etc. \Ac{EV} dataset from \cite{en14123692} consists of driving data of 5 identical vehicles over a year where each vehicle's data for charging and discharging events is sampled at 10-second intervals. An LPF battery dataset from \cite{RN713} is used for \acs{EV}'s \ac{SoH} estimation. It consists of 4 cells with a number of charge-discharge cycles over a thousand times for each cell. 

\textit{Layer 2:}
For the \ac{EV} layer in \cref{table:compare_dataset}, data from EA Technology \cite{web_dataset_Ref27} consists of residential \ac{EV} charging events with charging behaviors of over 200 participants observed from February 2014 to November 2015. EA Technology offers specialized asset management solutions for electrical asset owners and operators worldwide \cite{web_dataset_Ref27}. This data is used by \cite{Ref27} for the study of \ac{EV} charging behaviors. The iHomeLab PART dataset \cite{huber2020residential_R35r27} includes the energy consumption of households and specific appliances as well as \ac{PV} generation. The data are collected from 5 houses in the Lucerne region, Switzerland, and recorded over 1.5 to 3.5 years. It is used by \cite{Ref35} for data augmentation in the smart home applications, and by \cite{Ref45} for the study of adversarial attacks. \Ac{CLNR}'s dataset TC1a \cite{web_data_R43r1} and TC5 \cite{web_data_R43r2} consist of UK electricity customers’ overall electricity consumption and customers’ energy use and solar \ac{PV} performance, respectively. These datasets are used in \ac{GAN} \cite{Ref43} for generating residential \ac{EV} load. \Ac{CLNR} is a project supported by the Ofgem's Low Carbon Network Fund, which aimed to facilitate UK's low carbon energy sector \cite{web_Ref46_aboutCLNR}. 

\textit{Layer 3:}
This layer is about the interaction between \ac{EV}s and power grid. The City of Boulder Open Data Hub \cite{web_colorado_OpenData_gov} is used by a Transformer-based model for \ac{EV} load forecasting. It provides \ac{EV} charging load records spanning about 4 years collected from 25 public charging stations in Boulder, Colorado where the charging stations are equipped with 22 kW-rated connectors. Datasets from \cite{web_elia_data} and \cite{CaltechParkData_R28r38} are employed for scenario generations. The Belgian grid dataset from Elia Group \cite{web_elia_data} includes \ac{PV} and load in the microgrid with about two months of data. \Ac{ACN} data \cite{CaltechParkData_R28r38} include the time of \ac{EV} connection, done charging time, amount of energy received by \ac{EV}, and time of \ac{EV} leaving the parking lot. This dataset contains charging sessions at Caltech parking lots in California, and is continuously growing every day \cite{CaltechParkData_R28r38}. 

\textit{Layer 4:}
For the security layer, California \ac{ISO} \ac{OASIS} site \cite{web_CaliforniaISO} provides the continuously updating dataset for electricity prices and the grid load information where \ac{OASIS} offers real-time data related to the \ac{ISO} transmission system and its market. Data from \cite{web_CaliforniaISO} is used by \acs{RL-AdvGAN} \cite{Ref36} for the study of adversarial attacks. The load profiles from the grid can be found in \ac{ERCOT} \cite{web_ercot_data}. However, it requires an IP address from the U.S. to access data from \ac{ERCOT}. This dataset is used by \ac{GAE} \cite{Ref34} for the investigation of \ac{FDIAs}. \Ac{OTIDS} dataset \cite{web_Ref47r21_OTIDS_dataset} and Car-Hacking dataset \cite{web_Ref47r10_CarHacking_dataset} are employed in \cite{Ref47} for the detection of \ac{DoS}, fuzzy, and impersonation attacks. 

\textit{Multiple Layers:} 
Moreover, datasets can be used in multiple layers. For example, the Shanghai New Energy \ac{EV} Monitoring Center \cite{web_Shanghai_data} is a data sharing and cooperation platform with extensive operational data on new energy vehicles in Shanghai, and provides the \acs{EV}'s data such as speed, acceleration, battery status, and location. The data is continuously updated and used in \cite{Ref29} for the prediction of \ac{EV}'s \ac{SoC} and load.

\section{Future Directions}
\label{sect5:Future_Directions}

\Ac{GenAI} has demonstrated its potential in the \ac{IoEV}, and there are still areas to be improved. This section outlines several directions to unlock new opportunities in the \ac{IoEV} ecosystem.

\subsection{Improving Existing Solutions}
While \ac{GenAI} models have been developed for electric mobility, they are not perfect. 

\subsubsection{Up-to-date Models}
One challenge is to make sure models are updated and data plays an important role. Data patterns may evolve over time due to the dynamic nature of \ac{IoEV} environments. This requires the models to have continuous learning capability where new information should be adapted without forgetting previous knowledge. Integrating continual learning into the \ac{GenAI} system can keep it remain effective over the lifespan of an \ac{EV}. 

\subsubsection{Hallucination}
Another challenge could be solving the hallucination issues in \ac{GenAI} models. The hallucination refers to \ac{ML} generating outputs that are plausible but incorrect. For future work, developing methods that can detect and mitigate hallucinations in \ac{GenAI} is very important, especially for high-stakes applications such as \ac{EV} routing and EV's battery management system. The hybrid systems combining \ac{GenAI} with the traditional rule-based approach or the new architectures that can verify the correctness of generated content/data could be developed to avoid hallucination issues. 

\subsubsection{Transfer Learning}
Additionally, models, though are accurate for certain applications, may suffer from performance drops when the applications are different, even slightly. Transfer learning could be used to apply knowledge gained from one domain (e.g., load forecasting in Europe) to another (e.g., load forecasting in Asia), which reduces training time, lowers computational costs, and improves results with smaller datasets.

\subsubsection{Integration of EVs as Distributed Energy Resources}
Furthermore, combining traditional \ac{G2V} and advanced \ac{V2G} techniques allows \ac{EVs} act as mobile energy storage units and enables bidirectional energy flow between the vehicle and the grid. The \ac{G2V} techniques view \ac{EVs} as the energy consumers, while \ac{V2G} techniques provide \ac{EVs} with the opportunity to deliver power back to the grid. This bidirectional energy flow capability transforms \ac{EVs} into essential elements of distributed energy storage systems, provides the benefits to the grid such as helping for load balance, mitigating peak demand, enhancing grid resilience and stability. For future work, developing \ac{GenAI}-based optimization algorithms which can dynamically manage energy flow between \ac{EVs} and the grid, as well as energy transactions between \ac{EVs}, in coordination with renewable energy sources, is crucial for future of smart grids and sustainable energy solutions.

\subsection{New Methodologies}
Besides enhancing existing models, new technologies can also be explored.

\subsubsection{LLMs}
\acs{LLMs} are often discussed together with \ac{GenAI}, both representing the latest \ac{ML} technology advancements. \acs{LLMs} were originally developed for language-centric applications and language has not been the focus of \ac{EV} related applications. With the ever-increasing interactions between \ac{EV} systems and users, there is a demand to incorporate \ac{NLP} into the technology stack to facilitate the interaction and contribute to the enhancement of \ac{IoEV}. For example, \acs{LLMs} such as GPT-4 and LLaMA can offer enhanced interactions between \ac{EV} users and charging infrastructure. The \ac{EV} users can specify their charging preferences in natural language, such as \enquote{\textit{I need to charge my car fully by 7 AM, tomorrow morning}}. Then, the \ac{LLMs}-based system can translate this into an optimized charging schedule. Such interaction simplifies the process and offers a more accessible and personalized experience for the \ac{EV} users, not necessarily from a technical background of \ac{IoEV}. \ac{LLMs} can also be used to interpret users' historical queries, largely language-based, to predict user demand and optimize real-time responses. Overall, the integration of \ac{LLMs} in \ac{IoEV} charging systems shall be investigated to improve user experience and facilitate the adoption of electric mobility.

\subsubsection{Federated Learning} 
Privacy concerns, systems' robustness, and decentralized systems are critical in \ac{IoEV} applications. Federated learning can address security issues by enabling \ac{EVs} to learn collaboratively without sharing sensitive data. Future work could study how federated learning can be effectively implemented in large-scale \ac{IoEV} to promote the system's scalability and security. 

\subsubsection{Hybrid Models}
When one model is insufficient to perform well, the hybrid models can be considered and the models are not limited to GenAI. For example, combining \ac{GenAI} with \ac{DRL} may improve the models' performance in complex \ac{IoEV} applications. Several existing research efforts are \ac{GRU} and \ac{VAE} \cite{Ref20} for battery anomaly detection, \ac{LSTM} and \ac{GAN} for \ac{SoC} estimation \cite{Ref42}, \ac{DNN} and \ac{GAN} for EV charging behaviors prediction \cite{Ref27}, etc. Future research may explore cross-layer optimizations with hybrid models, e.g., advanced \ac{BMS} informs routing decisions and grid interactions. This helps create an integrated and efficient system.

\subsubsection{Embodied AI and AGI}
Looking toward future advancements, the embodied AI and AGI are expected to play transformative roles in IoEV. The former combines cognitive processing with physical interaction to enhance adaptability and real-world decision-making, e.g., navigating complex environments and responding to traffic conditions. The latter aims at achieving broad cognitive abilities similar to human intelligence. It can potentially revolutionize IoEV, e.g., make decisions in unpredictable environments without relying on pre-programmed instructions. 

Together, new methodologies and technologies could drive further innovation in IoEV in various aspects such as sustainability and safety.

\section{Conclusions}
\label{sect6:Conclusions}

In this survey, we explored the applications of \ac{GenAI} in the \ac{IoEV} from various perspectives. We grouped relevant applications across four different layers: EV's battery layer, individual \ac{EV} layer, the grid layer, and security layer. We concluded that these applications primarily utilize GenAI's capabilities in data feature extraction, enhancement, and generation. The characteristics of \ac{GenAI} make it ideal for data augmentation. Hence, research works in the first three layers all used \ac{GenAI} for this purpose. At layer 1, \acs{GenAIs} including \ac{GAN}, \ac{GDM}, \ac{VAE}, and Transformer were introduced for anomaly detection, as well as \ac{SoC} and \ac{SoH} estimations of the EV's battery. At layer 2, \ac{GAN} and Transformer were discussed for \ac{EV} charging behaviors and loads as well as the optimal \ac{EV} routing problem. At layer 3, \ac{GAN}, \ac{VAE}, and Transformer are the main techniques currently employed for \ac{EV} charging load forecasting/charging scenarios generation, and \ac{LLMs} for the analysis of \ac{EV} charging reviews. At layer 4, \ac{GAN} and \ac{AE} were often applied for the detection and generation of attacks that may threaten the systems across layers 1 to 3. Subsequently, we summarized the publicly available dataset and the possible further research directions for GenAI's applications in \ac{IoEV}. In conclusion, this survey highlights the essential role of \ac{GenAI} in \ac{IoEV} and underscores the urgent need for further exploration of its applications.

\balance

\bibliographystyle{IEEEtran}
\bibliography{Refbib}

\newpage

\end{document}

%% file: abbr.tex
\DeclareAcronym{GenAI}{
	short = GenAI,
	long  = generative artificial intelligence,
	class = abbrev
}

\DeclareAcronym{LRA}{
	short = LRA,
	long  = low-rank tensor approximation,
	class = abbrev
}

\DeclareAcronym{SoH}{
	short = SoH,
	long  = state of health,
	class = abbrev
}

\DeclareAcronym{V2G}{
	short = V2G,
	long  = vehicle-to-grid,
	class = abbrev
}

\DeclareAcronym{G2V}{
	short = G2V,
	long  = grid-to-vehicle,
	class = abbrev
}

\DeclareAcronym{LSTM}{
	short = LSTM,
	long  = long short-term memory,
	class = abbrev
}

\DeclareAcronym{SAC}{
	short = SAC,
	long  = soft actor-critic,
	class = abbrev
}

\DeclareAcronym{PPO}{
	short = PPO,
	long  = proximal policy optimization,
	class = abbrev
}

\DeclareAcronym{LLMs}{
	short = LLMs,
	long  = large language models,
	class = abbrev
}

\DeclareAcronym{HEVs}{
	short = HEVs,
	long  = hybrid electric vehicles,
	class = abbrev
}

\DeclareAcronym{HEV}{
	short = HEV,
	long  = hybrid electric vehicle,
	class = abbrev
}

\DeclareAcronym{RL}{
	short = RL,
	long  = reinforcement learning,
	class = abbrev
}

\DeclareAcronym{MADRL}{
	short = MADRL,
	long  = multi-agent deep reinforcement learning,
	class = abbrev
}

\DeclareAcronym{MARL}{
	short = MARL,
	long  = multi-agent reinforcement learning,
	class = abbrev
}

\DeclareAcronym{RDNs}{
	short = RDNs,
	long  = radial distribution networks,
	class = abbrev
}

\DeclareAcronym{SDRL}{
	short = SDRL,
	long  = safe deep reinforcement learning,
	class = abbrev
}

\DeclareAcronym{CMDP}{
	short = CMDP,
	long  = constrained Markov decision process,
	class = abbrev
}

\DeclareAcronym{MDP}{
	short = MDP,
	long  = Markov decision process,
	class = abbrev
}

\DeclareAcronym{CDDPG}{
	short = CDDPG,
	long  = charging control deep deterministic policy gradient,
	class = abbrev
}

\DeclareAcronym{DDPG}{
	short = DDPG,
	long  = deep deterministic policy gradient,
	class = abbrev
}

\DeclareAcronym{OPF}{
	short = OPF,
	long  = optimal power flow,
	class = abbrev
}

\DeclareAcronym{ILP}{
	short = ILP,
	long  = integer linear programming,
	class = abbrev
}

\DeclareAcronym{EVCS}{
	short = EVCS,
	long  = electric vehicle charging station,
	class = abbrev
}

\DeclareAcronym{MASCO}{
	short = MASCO,
	long  = multiagent selfish-collaborative,
	class = abbrev
}

\DeclareAcronym{LDF}{
	short = LDF,
	long  = less demand first,
	class = abbrev
}

\DeclareAcronym{SMADRL}{
	short = SMADRL,
	long  = Stackelberg multi-agent deep reinforcement learning,
	class = abbrev
}

\DeclareAcronym{AePPO}{
	short = AePPO,
	long  = adaptive exploration proximal policy optimization,
	class = abbrev
}

\DeclareAcronym{SVQR}{
	short = SVQR,
	long  = support vector quantile regression,
	class = abbrev
}

\DeclareAcronym{QR}{
	short = QR,
	long  = linear quantile regression,
	class = abbrev
}

\DeclareAcronym{GBQR}{
	short = GBQR,
	long  = gradient boosting quantile regression,
	class = abbrev
}

\DeclareAcronym{DQN}{
	short = DQN,
	long  = deep Q-networks,
	class = abbrev
}

\DeclareAcronym{GAN}{
	short = GAN,
	long  = generative adversarial network,
	class = abbrev
}
 
\DeclareAcronym{SMOTE}{
	short = SMOTE,
	long  = synthetic minority oversampling technique,
	class = abbrev
}

\DeclareAcronym{GMM}{
	short = GMM,
	long  = Gaussian mixture model,
	class = abbrev
}

\DeclareAcronym{GRU}{
	short = GRU,
	long  = gated recurrent unit,
	class = abbrev
}

\DeclareAcronym{RNN}{
	short = RNN,
	long  = recurrent neural network,
	class = abbrev
}

\DeclareAcronym{MRNN}{
	short = MRNN,
	long  = modular recurrent neural network,
	class = abbrev
}

\DeclareAcronym{SVR}{
	short = SVR,
	long  = support vector regression,
	class = abbrev
}

\DeclareAcronym{RF}{
	short = RF,
	long  = random forest,
	class = abbrev
}

\DeclareAcronym{VAE}{
	short = VAE,
	long  = variational autoencoder,
	class = abbrev
}

\DeclareAcronym{CAN}{
	short = CAN,
	long  = controller area network,
	class = abbrev
}

\DeclareAcronym{ECUs}{
	short = ECUs,
	long  = Electronic Control Units,
	class = abbrev
}

\DeclareAcronym{MVTS}{
	short = MVTS,
	long  = multivariate time series,
	class = abbrev
}

\DeclareAcronym{EVRP}{
	short = EVRP,
	long  = electric vehicle routing problem,
	class = abbrev
}

\DeclareAcronym{VRP}{
	short = VRP,
	long  = vehicle routing problem,
	class = abbrev
}

\DeclareAcronym{NLP}{
	short = NLP,
	long  = natural language processing,
	class = abbrev
}

\DeclareAcronym{GAIL}{
	short = GAIL,
	long  = generative adversarial imitation learning,
	class = abbrev
}

\DeclareAcronym{GAIN}{
	short = GAIN,
	long  = generative adversarial imputation network,
	class = abbrev
}

\DeclareAcronym{SoC}{
	short = SoC,
	long  = state of charge,
	class = abbrev
}

\DeclareAcronym{MAPPO}{
	short = MAPPO,
	long  = multi-agent proximal policy optimization,
	class = abbrev
}

\DeclareAcronym{FDIAs}{
	short = FDIAs,
	long  = false data injection attacks,
	class = abbrev
}

\DeclareAcronym{IoEV}{
	short = IoEV,
	long  = Internet of electric vehicles,
	class = abbrev
}

\DeclareAcronym{IoV}{
	short = IoV,
	long  = Internet of vehicle,
	class = abbrev
}

\DeclareAcronym{V2V}{
	short = V2V,
	long  = vehicle-to-vehicle,
	class = abbrev
}

\DeclareAcronym{V2I}{
	short = V2I,
	long  = vehicle-to-infrastructure,
	class = abbrev
}

\DeclareAcronym{V2X}{
	short = V2X,
	long  = vehicle-to-everything,
	class = abbrev
}

\DeclareAcronym{ADMM}{
	short = ADMM,
	long  = alternating direction method of multipliers,
	class = abbrev
}

\DeclareAcronym{MCVs}{
	short = MCVs,
	long  = mobile charging vehicles,
	class = abbrev
}

\DeclareAcronym{IoT}{
	short = IoT,
	long  = Internet of things,
	class = abbrev
}

\DeclareAcronym{SVM}{
	short = SVM,
	long  = support vector machine,
	class = abbrev
}

\DeclareAcronym{GPR}{
	short = GPR,
	long  = Gaussian process regression,
	class = abbrev
}

\DeclareAcronym{ANN}{
	short = ANN,
	long  = artificial neural network,
	class = abbrev
}

\DeclareAcronym{NMG}{
	short = NMG,
	long  = networked microgrids,
	class = abbrev
}

\DeclareAcronym{MLLMs}{
	short = MLLMs,
	long  = multimodal large language models,
	class = abbrev
}

\DeclareAcronym{HDB}{
	short = HDB,
	long  = housing and development board,
	class = abbrev
}

\DeclareAcronym{UAVs}{
	short = UAVs,
	long  = unmanned aerial vehicles,
	class = abbrev
}

\DeclareAcronym{CAGR}{
	short = CAGR,
	long  = compound annual growth rate,
	class = abbrev
}

\DeclareAcronym{TOU}{
	short = TOU,
	long  = time-of-use,
	class = abbrev
}

\DeclareAcronym{TD3}{
	short = TD3,
	long  = twin delayed deep deterministic policy gradient,
	class = abbrev
}

\DeclareAcronym{DDPM}{
	short = DDPM,
	long  = denoising diffusion probabilistic model,
	class = abbrev
}

\DeclareAcronym{DDIM}{
	short = DDIM,
	long  = denoising diffusion implicit model,
	class = abbrev
}

\DeclareAcronym{BMS}{
	short = BMS,
	long  = battery management system,
	class = abbrev
}

\DeclareAcronym{MD-MTD}{
	short = MD-MTD,
	long  = multi-distribution global trend diffusion,
	class = abbrev
}

\DeclareAcronym{AutoML}{
	short = AutoML,
	long  = automated machine learning,
	class = abbrev
}

\DeclareAcronym{KNNR}{
	short = KNNR,
	long  = $k$-nearest neighbor regressor,
	class = abbrev
}

\DeclareAcronym{SMAPE}{
	short = SMAPE,
	long  = symmetric mean absolute percentage error,
	class = abbrev
}

\DeclareAcronym{AE-GAN}{
	short = AE-GAN,
	long  = Autoencoder with GANs,
	class = abbrev
}

\DeclareAcronym{EMS}{
	short = EMS,
	long  = energy management system,
	class = abbrev
}

\DeclareAcronym{ESS}{
	short = ESS,
	long  = energy storage system,
	class = abbrev
}

\DeclareAcronym{EM-EVRP}{
	short = EM-EVRP,
	long  = energy-minimization EV routing problem,
	class = abbrev
}

\DeclareAcronym{ACO}{
	short = ACO,
	long  = ant colony algorithm ,
	class = abbrev
}

\DeclareAcronym{ALNS}{
	short = ALNS,
	long  = adaptive large neighborhood search,
	class = abbrev
}

\DeclareAcronym{AM}{
	short = AM,
	long  = DRL with general attention model,
	class = abbrev
}

\DeclareAcronym{DM-EVRP}{
	short = DM-EVRP,
	long  = distance minimization electric vehicle routing problem,
	class = abbrev
}

\DeclareAcronym{FGSM}{
	short = FGSM,
	long  = fast gradient sign method,
	class = abbrev
}

\DeclareAcronym{BIM}{
	short = BIM,
	long  = basic iterative method,
	class = abbrev
}

\DeclareAcronym{ARMA}{
	short = ARMA,
	long  = autoregressive moving average,
	class = abbrev
}

\DeclareAcronym{ARIMA}{
	short = ARIMA,
	long  = autoregressive integral moving average,
	class = abbrev
}

\DeclareAcronym{SSMs}{
	short = SSMs,
	long  = state space models,
	class = abbrev
}

\DeclareAcronym{NCM}{
	short = NCM,
	long  = nickel cobalt manganese,
	class = abbrev
}

\DeclareAcronym{SARIMA}{
	short = SARIMA,
	long  = seasonal autoregressive integrated moving average,
	class = abbrev
}

\DeclareAcronym{SSIM}{
	short = SSIM,
	long  = structured similarity indexing method,
	class = abbrev
}

\DeclareAcronym{FSIM}{
	short = FSIM,
	long  = feature similarity indexing method,
	class = abbrev
}

\DeclareAcronym{ACN}{
	short = ACN,
	long  = adaptive charging network,
	class = abbrev
}

\DeclareAcronym{OASIS}{
	short = OASIS,
	long  = open access same-time information system,
	class = abbrev
}

\DeclareAcronym{ISO}{
	short = ISO,
	long  = independent system operator,
	class = abbrev
}

\DeclareAcronym{DBN}{
	short = DBN,
	long  = deep belief network,
	class = abbrev
}

\DeclareAcronym{ELM}{
	short = ELM,
	long  = extreme learning machine,
	class = abbrev
}

\DeclareAcronym{GAE}{
	short = GAE,
	long  = graph autoencoder,
	class = abbrev
}

\DeclareAcronym{GHG}{
	short = GHG,
	long  = global greenhouse gas,
	class = abbrev
}

\DeclareAcronym{HEDGE}{
	short = HEDGE,
	long  = home electricity data generator,
	class = abbrev
}

\DeclareAcronym{LOF}{
	short = LOF,
	long  = local outlier factor,
	class = abbrev
}

\DeclareAcronym{OCSVM}{
	short = OCSVM,
	long  = one-class support vector machines,
	class = abbrev
}

\DeclareAcronym{PCA}{
	short = PCA,
	long  = principal component analysis,
	class = abbrev
}

\DeclareAcronym{V2M}{
	short = V2M,
	long  = vehicle-to-microgrid,
	class = abbrev
}

\DeclareAcronym{DBSCAN}{
	short = DBSCAN,
	long  = density-based spatial clustering of applications with noise,
	class = abbrev
}

\DeclareAcronym{BESS}{
	short = BESS,
	long  = battery energy storage system,
	class = abbrev
}

\DeclareAcronym{RX-ADS}{
	short = RX-ADS,
	long  = interpretable anomaly detection system,
	class = abbrev
}

\DeclareAcronym{T5}{
	short = T5,
	long  = text-to-text transfer Transformer,
	class = abbrev
}

\DeclareAcronym{IS}{
	short = IS,
	long  = inception score,
	class = abbrev
}

\DeclareAcronym{FID}{
	short = FID,
	long  = Fréchet inception distance,
	class = abbrev
}

\DeclareAcronym{GDM}{
	short = GDM,
	long  = generative diffusion model,
	class = abbrev
}

\DeclareAcronym{DRL}{
	short = DRL,
	long  = deep reinforcement learning,
	class = abbrev
}

\DeclareAcronym{EV}{
	short = EV,
	long  = electric vehicle,
	class = abbrev
}

\DeclareAcronym{AE}{
	short = AE,
	long  = Autoencoder,
	class = abbrev
}

\DeclareAcronym{AI}{
	short = AI,
	long  = artificial intelligence,
	class = abbrev
}

\DeclareAcronym{CNN}{
	short = CNN,
	long  = convolutional neural network,
	class = abbrev
}

\DeclareAcronym{DNN}{
	short = DNN,
	long  = deep neural network,
	class = abbrev
}

\DeclareAcronym{FNN}{
	short = FNN,
	long  = feed-forward neural network,
	class = abbrev
}

\DeclareAcronym{DoS}{
	short = DoS,
	long  = denial of service,
	class = abbrev
}

\DeclareAcronym{PV}{
	short = PV,
	long  = photovoltaic,
	class = abbrev
}

\DeclareAcronym{KNN}{
	short = KNN,
	long  = $k$-nearest neighbor,
	class = abbrev
}

\DeclareAcronym{BERTtery}{
	short = BERTtery,
	long  = bidirectional encoder representations from Transformers for batteries,
	class = abbrev
}

\DeclareAcronym{BERT}{
	short = BERT,
	long  = bidirectional encoder representations from Transformers,
	class = abbrev
}
 
\DeclareAcronym{GAN-RF}{
	short = GAN-RF,
	long  = GAN model with random forest,
	class = abbrev
}

\DeclareAcronym{L-BFGS}{
	short = L-BFGS,
	long  = limited-memory Broyden-Fletcher-Goldfarb-Shanno,
	class = abbrev
}

\DeclareAcronym{CopulaGAN}{
	short = CopulaGAN,
	long  = Copula generative adversarial network,
	class = abbrev
}

\DeclareAcronym{CR-GAN}{
	short = CR-GAN,
	long  = convolutional recursive GAN,
	class = abbrev
}

\DeclareAcronym{RMSE}{
	short = RMSE,
	long  = root mean square error,
	class = abbrev
}

\DeclareAcronym{SC-WGAN-GP}{
	short = SC-WGAN-GP,
	long  = similarity-constrained Wasserstein GAN with gradient penalty,
	class = abbrev
}

\DeclareAcronym{TimeGAN}{
	short = TimeGAN,
	long  = time-series GAN,
	class = abbrev
}

\DeclareAcronym{TS-DCGAN}{
	short = TS-DCGAN,
	long  = time-series deep convolutional GAN,
	class = abbrev
}

\DeclareAcronym{TS-WGAN}{
	short = TS-WGAN,
	long  = time-series Wasserstein GAN,
	class = abbrev
}

\DeclareAcronym{WGAN-GP}{
	short = WGAN-GP,
	long  = Wasserstein GAN with gradient penalty,
	class = abbrev
}

\DeclareAcronym{WGAN}{
	short = WGAN,
	long  = Wasserstein GAN,
	class = abbrev
}

\DeclareAcronym{DCGAN}{
	short = DCGAN,
	long  = deep convolutional GAN,
	class = abbrev
}

\DeclareAcronym{CGAN}{
	short = CGAN,
	long  = conditional generative adversarial network,
	class = abbrev
}

\DeclareAcronym{RC-GAN}{
	short = RC-GAN,
	long  = recurrent conditional GAN,
	class = abbrev
}

\DeclareAcronym{MAE}{
	short = MAE,
	long  = mean absolute error,
	class = abbrev
}

\DeclareAcronym{ML}{
	short = ML,
	long  = machine learning,
	class = abbrev
}

\DeclareAcronym{AGI}{
	short = AGI,
	long  = artificial general intelligence,
	class = abbrev
}

\DeclareAcronym{BNN}{
	short = BNN,
	long  = Bayesian neural network,
	class = abbrev
}

\DeclareAcronym{KL}{
	short = KL,
	long  = Kullback-Leibler,
	class = excluded
}

\DeclareAcronym{CW-GAN}{
	short = CW-GAN,
	long  = convolution conditional GANs with Wasserstein distance as network objective function,
	class = excluded
}

\DeclareAcronym{DSO}{
	short = DSO,
	long  = distribution system operator,
	class = excluded
}

\DeclareAcronym{CSOs}{
	short = CSOs,
	long  = charging station operators,
	class = excluded
}

\DeclareAcronym{ICEs}{
	short = ICEs,
	long  = internal combustion engines,
	class = excluded
}

\DeclareAcronym{GRU-VAE}{
	short = GRU-VAE,
	long  = gated recurrent unit based variational autoencoder,
	class = excluded
}

\DeclareAcronym{RL-AdvGAN}{
	short = RL-AdvGAN,
	long  = a GAN-based adversarial attack against DRL algorithms,
	class = excluded
}

\DeclareAcronym{VAE-GAN}{
	short = VAE-GAN,
	long  = variational autoencoder-generative adversarial network,
	class = excluded
}

\DeclareAcronym{GRU-GAN}{
	short = GRU-GAN,
	long  = GAN model with GRU-based discriminator and generator,
	class = excluded
}

\DeclareAcronym{GDMs}{
	short = GDMs,
	long  = generative diffusion models,
	class = excluded
}

\DeclareAcronym{MAPE}{
	short = MAPE,
	long  = mean absolute percentage error,
	class = excluded
}

\DeclareAcronym{Phy-ADS}{
	short = Phy-ADS,
	long  = physical anomaly detection ,
	class = excluded
}

\DeclareAcronym{Cy-ADS}{
	short = Cy-ADS,
	long  = cyber anomaly detection,
	class = excluded
}

\DeclareAcronym{RPM}{
	short = RPM,
	long  = revolutions per minute,
	class = excluded
}

\DeclareAcronym{CLNR}{
	short = CLNR,
	long  = customer-led network revolution,
         class = excluded       
}

\DeclareAcronym{OTIDS}{
	short = OTIDS,
	long  = offset ratio and time interval based intrusion detection system,
	class = excluded
}

\DeclareAcronym{HCRL}{
	short = HCRL,
	long  = hacking and countermeasures research lab,
	class = excluded
}

\DeclareAcronym{ERCOT}{
	short = ERCOT,
	long  = electric reliability council of Texas,
	class = excluded
}

\DeclareAcronym{NSMC-EV}{
	short = NSMC-EV,
	long  = National Service and Management Center for Electric Vehicles,
	class = excluded
}

\DeclareAcronym{C-LSTM-WGAN-GP}{
	short = C-LSTM-WGAN-GP,
	long  = conditional LSTM Wasserstein GAN with gradient penalty,
         class = excluded       
}

\DeclareAcronym{C-LSTM-GAN}{
	short = C-LSTM-GAN,
	long  = conditional LSTM GAN,
	class = excluded
}

\DeclareAcronym{C-LSTM-WGAN}{
	short = C-LSTM-WGAN,
	long  = conditional LSTM Wasserstein GAN,
	class = excluded
}

\DeclareAcronym{CNNs}{
	short = CNNs,
	long  = convolutional neural networks,
	class = excluded
}

\DeclareAcronym{DNNs}{
	short = DNNs,
	long  = deep neural networks,
	class = excluded
}

\DeclareAcronym{RNNs}{
	short = RNNs,
	long  = recurrent neural networks,
	class = excluded
}

\DeclareAcronym{EVs}{
	short = EVs,
	long  = electric vehicles,
	class = excluded
}

\DeclareAcronym{GenAIs}{
	short = GenAIs, 
	long  = generative artificial intelligences,
	class = excluded 
} 

\DeclareAcronym{VAEs}{
	short = VAEs,
	long  = variational autoencoders,
	class = excluded
}

\DeclareAcronym{GANs}{
	short = GANs,
	long  = generative adversarial networks,
	class = excluded 
}

\DeclareAcronym{EMAmath}{
	short = $\bar{p}_{ema}$,
	long  = {exponential moving average of the real-time price},
	sort  = $\bar{p}_{ema}$,
	class = nomencl
}